\documentclass{article}

\PassOptionsToPackage{numbers, compress}{natbib}

\usepackage[main, final]{neurips_2026}

\usepackage[utf8]{inputenc} %
\usepackage[T1]{fontenc}    %
\usepackage{hyperref}       %
\usepackage{url}            %
\usepackage{booktabs}       %
\usepackage{amsfonts}       %
\usepackage{nicefrac}       %
\usepackage{microtype}      %
\usepackage{xcolor}         %

\definecolor{mygreen}{HTML}{4FB283}

\usepackage[table]{xcolor}
\usepackage{booktabs}
\usepackage{makecell}

\usepackage{graphicx}
\usepackage{amsmath}
\usepackage{natbib}
\usepackage{multirow}

\usepackage{threeparttable}
\usepackage{float}
\usepackage{wrapfig}
\floatstyle{ruled}
\usepackage{algorithm}
\usepackage{algorithmic}

\usepackage{chngcntr}
\usepackage{etoc}
\usepackage{titlesec}

\usepackage{amsmath}
\usepackage{amssymb}
\usepackage{amsthm}

\usepackage{tabularx}
\usepackage{array}

\usepackage{caption}
\newcolumntype{C}{>{\centering\arraybackslash}X}

\graphicspath{{figures/}}

\title{Temporal Selective Exploration for Reinforcement Learning-Guided Continuous-Discrete Flow Matching in 3D Molecular Design}

\author{%
  \textbf{Lianghong Chen$^{1}$, Yan Yi Li$^{2}$, Gen Zhou$^{1}$, Yuxi Long$^{1}$,}\\
  \textbf{Ganlin Feng$^{1}$, Mike Domaratzki$^{1}$, Pingzhao Hu$^{1,2}$\thanks{Corresponding author. Department of Computer Science, Western University, 1400 Western Road, London, Ontario N6G 2V4, Canada. Email: \texttt{phu49@uwo.ca} (P.H.)}}\\
  $^{1}$Department of Computer Science, Western University, London, Ontario, Canada\\
  $^{2}$Department of Biochemistry, Western University, London, Ontario, Canada
}

\begin{document}

\maketitle

\begin{abstract}

Flow matching has shown strong potential in generative tasks, while its optimization via Reinforcement Learning (RL) remains underexplored, especially in continuous-discrete mixed settings. Moreover, flow matching is formulated as an Ordinary Differential Equation (ODE), whose deterministic trajectories do not naturally support RL optimization. Introducing stochasticity by converting the entire trajectory into a Stochastic Differential Equation (SDE) is a common practice. Nevertheless, not all timesteps contribute to the final outcomes, and redundant exploration introduces additional noise. Furthermore, in multi-objective settings, some objectives may dominate the optimization while others receive insufficient updates. In this study, we propose an RL-guided continuous-discrete flow matching framework with temporal selective exploration for 3D de novo molecular design. Specifically, our method jointly optimizes continuous and discrete flow matching for atomic coordinates and molecular identities, respectively, while restricting exploration to timesteps that primarily contribute to target properties, reducing ineffective exploration. We also introduce a reward-based advantage calibration mechanism to mitigate objective dominance in multi-objective optimization. The framework is applied to design inhibitors for both the well-studied target protein EGFR and the challenging Pin1 protein. We identify some novel candidate inhibitors, supported by in silico validation.
\end{abstract}

\section{Introduction}

\textbf{Background.} Many aggressive cancers, such as Triple-Negative Breast Cancer (TNBC)~\cite{TNBC}, still lack effective therapies~\cite{Cancer-1,Cancer-2}. Discovering new drugs may provide potential therapeutic strategies for these diseases~\cite{DrugDiscovery-1,DrugDiscovery-2,DrugDiscovery-3}. The progression of many aggressive cancers is closely associated with dysregulated protein activity~\cite{Cancer-3}. For example, TNBC has been linked to the overexpression and aberrant activation of Peptidyl-prolyl cis-trans isomerase NIMA-interacting 1 (Pin1)~\cite{TNBC-pin1}. To modulate these activities, a common strategy is to develop inhibitors targeting these proteins~\cite{DevelopInhibitors}. These inhibitors are typically small molecules~\cite{small-molecules}. During their lead discovery stage, molecules are expected to exhibit favorable drug-likeness, synthetic accessibility, and strong binding affinity to the target protein~\cite{need-properties}. These properties are inherently tied to molecular structures~\cite{tie-to-structure}. For instance, drug-likeness and synthetic accessibility are related to functional groups and bonding patterns~\cite{group-and-bond-patt}, while binding affinity is governed by the 3D arrangement of atoms~\cite{3D-arrangement}. Therefore, to jointly account for these properties, explicitly modeling molecules in 3D has become a widely adopted paradigm~\cite{G-SchNet,E-NF,EDM,GeoLDM,JODO}.

\textbf{Challenge 1: RL optimization for continuous-discrete flow matching.} Recently, flow matching models have shown strong performance in 3D de novo molecular generation, especially in continuous-discrete mixed frameworks~\cite{FMisPromising_paradigm,FlowMol-CTMC,FlowMol3}. However, such frameworks have not been fully applied to practical drug discovery tasks, as they lack consideration of key drug-relevant properties. To guide the model to generate molecules with desired properties, a common strategy is to use Reinforcement Learning (RL) to optimize generative models~\cite{RL4generative-DDPO,RL4generative-Survey}. In previous studies, RL has been widely used to optimize continuous frameworks, while RL for Continuous-Time Markov Chain (CTMC)~\cite{CTMC}-based discrete modeling, as well as for continuous-discrete frameworks, remains underexplored~\cite{UDM-GRPO}. %
The key challenge lies in defining valid likelihoods for stochastic discrete transitions within continuous-discrete flow matching frameworks. In CTMC-based molecular generation, discrete components evolve through stochastic masking and unmasking, while the continuous coordinates follow transport trajectories. As a result, RL optimization must account for both continuous and discrete transition likelihoods under the same generation process, since mismatches between the sampling dynamics and likelihood evaluation may introduce inconsistencies during optimization.

\begin{wrapfigure}{r}{0.49\textwidth}
\centering
\includegraphics[width=\linewidth,clip]{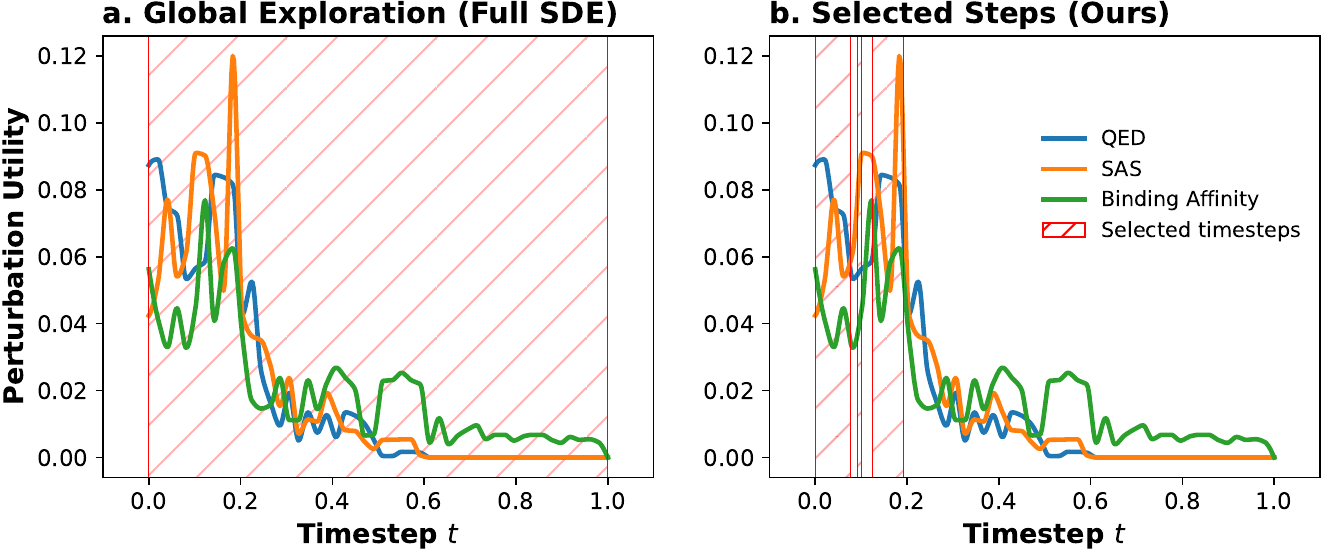}
\caption{Illustration of temporal selective exploration. The curves show the sensitivity of QED, SAS, and binding affinity to perturbations along the generation trajectory. The shaded regions indicate timesteps used for stochastic exploration under (a) Full-SDE and (b) temporal selection.}
\label{fig:overview_challenges}
\end{wrapfigure}

\textbf{Challenge 2: Timestep selection.} RL has been successfully used to optimize continuous flow matching models in image generation by introducing stochasticity into the originally deterministic Ordinary Differential Equation (ODE) trajectories~\cite{FlowMatching-ODE}, typically by converting the entire generation process into a Stochastic Differential Equation (SDE) formulation~\cite{FlowGRPO}. This stochastic formulation enables exploration and provides learning signals for RL by perturbing the sampling process~\cite{FlowGRPO}. However, such formulations typically assume that all timesteps contribute to the final outcome. In practice, flow matching dynamics are highly non-uniform over time, and different timesteps have varying influence on the resulting molecular structures and properties~\cite{non-uniformity,non-uniformity-2}. As a result, not all timesteps produce meaningful learning signals when stochasticity is introduced~\cite{non-uniformity-2}. Effective updates are often concentrated on a small subset of timesteps, while exploration at the remaining timesteps contributes little to performance improvement and may introduce additional noise~\cite{SDE-problem-1,SDE-problem-2,SDE-problem-3,SDE-problem-4}.

\textbf{Challenge 3: Multi-objective dominance.} In practical settings, optimizing generative models with RL often involves multiple objectives. For example, in molecular design, multiple properties need to be optimized simultaneously. However, these objectives are often inherently conflicting, and their reward signals can differ significantly in scale~\cite{conflicting_scale}, gradient sensitivity, and optimization difficulty~\cite{optimization_difficulty}. When reward gradients vary substantially, these differences can lead to imbalanced updates across objectives~\cite{gradnorm}. Easier objectives may tend to dominate the training process~\cite{MGDA}, while harder objectives fail to receive sufficient effective updates~\cite{hard_objectives}. As a result, the learning process becomes biased toward certain objectives, leading to inconsistent improvements.

\textbf{Overview.} In this study, we propose a multi-objective RL-guided continuous-discrete flow matching framework with temporal selective exploration (Fig.~\ref{fig:overview_challenges}) for 3D de novo molecular design. We perform RL optimization using unified continuous-discrete transition likelihoods, where likelihoods are evaluated under the same distribution that generates the transitions. For the continuous part, stochasticity is introduced via ODE-to-SDE conversion, transforming deterministic transport into a probabilistic process with tractable likelihood. The overall trajectory is defined as a joint distribution over continuous and discrete transitions, enabling RL optimization over both components. Additionally, by perturbing the generation process, we identify key timesteps that are informative for property improvement and restrict RL exploration and updates to these regions. We further introduce Cross-Reward Advantage Calibration (CRAC), which balances objectives based on their remaining gaps and joint satisfaction of target criteria. Our primary task is to design novel inhibitors with therapeutic potential for two cancer-related target proteins with different levels of design difficulty: the well-studied Epidermal Growth Factor Receptor (EGFR)~\cite{EGFR} and the more challenging Pin1~\cite{Pin1}. We focus on optimizing three key molecular properties, namely Quantitative Estimate of Drug-likeness (QED)~\cite{QED}, Synthetic Accessibility Score (SAS)~\cite{group-and-bond-patt}, and binding affinity~\cite{bindingaffinity}. Experiments are mainly conducted on the large-scale small-molecule dataset GEOM-Drugs. To validate the effectiveness of the proposed method, we compare against multiple baseline approaches and perform extensive ablation studies. For the top generated candidates, we further compare them with known ligands from protein-ligand co-crystal structures, and conduct Absorption, Distribution, Metabolism, Excretion, and Toxicity (ADMET)~\cite{ADMET-AI} analysis and Molecular Dynamics (MD)~\cite{MD} simulations to assess their drug-like properties and binding stability with the target proteins. To evaluate methodological generality, we further extend our experiments to an additional dataset, QM9. The results highlight the potential of the proposed framework for designing novel inhibitors and advancing automated molecular discovery.

\textbf{Our contributions can be summarized as follows:}
\begin{itemize}
\item We propose an RL-guided continuous-discrete flow matching framework for 3D de novo molecular design.

\item We introduce a temporal selective exploration strategy to reduce ineffective exploration during RL optimization.
	
\item We propose a CRAC mechanism to balance multiple objectives.

\item We identify novel candidate inhibitors for Pin1 and EGFR, supported by in silico evaluation.
\end{itemize}

\section{Related Works}

\textbf{Flow Matching for 3D Molecular Generation.}
Early works such as EquiFM \cite{EquiFM}, SemlaFlow \cite{SemlaFlow}, and FlowMol \cite{FlowMol} primarily adopt continuous flow matching, focusing on modeling molecular structures in continuous spaces, where both atomic coordinates and categorical variables are handled through continuous representations. Subsequently, FlowMol-CTMC \cite{FlowMol-CTMC} and FlowMol3 \cite{FlowMol3} explicitly introduce CTMC formulations to model the discrete components of molecules, such as atom types, bonds, and charges, enabling true discrete jump processes, while the continuous components (i.e., atomic coordinates) are still modeled via continuous flow matching. Among them, FlowMol3 achieves state-of-the-art performance. Most of these works focus on unconditional generation. PropMolFlow~\cite{PropMolFlow} extends this line of research by incorporating molecular properties into the generation process. However, it remains limited to continuous flow matching and is primarily designed for single-property generation. In addition, its conditioning mechanism limits control over the generation process and may hinder alignment with desired properties. Furthermore, as discussed in Appendix~\ref{tab:conditional_vs_unconditional}, conditional generation frameworks can be less flexible for downstream tasks and incur higher adaptation costs when extending to new objectives~\cite{unconditional-good}.

\textbf{Sampling and Optimization Strategies.}
Multiple strategies can be used to obtain more desirable samples from generative models. Sampling-based methods such as Best-of-N \cite{Best-of-n} and Sequential Monte Carlo (SMC) \cite{SMC} improve output quality by selecting or resampling candidates that better satisfy the objectives, without modifying the underlying model. While effective, these methods are computationally expensive and may impact the diversity of generated samples. Optimization-based methods typically rely on RL to guide models toward desired objectives, enabling systematic and scalable improvement. RL has been applied across various generative paradigms, including Variational AutoEncoders~\cite{VAE}, diffusion models \cite{diffusion}, and flow matching models \cite{FlowGRPO} in image generation. As a representative work, Flow-GRPO \cite{FlowGRPO} applies RL to flow matching in continuous domains by introducing stochasticity through ODE-to-SDE conversion across all timesteps, enabling policy optimization over the generation process. However, Flow-GRPO is designed for conditional image generation, where multiple samples can be generated under the same condition, enabling group-based optimization methods such as GRPO~\cite{GRPO}. In contrast, as discussed in Appendix~\ref{appendix:Choice_of_Optimization_Framework}, such approaches are less directly applicable to unconditional settings. Additionally, although prior works such as TR2-D2~\cite{TR2-D2} and UDM-GRPO~\cite{UDM-GRPO} attempt to optimize discrete generative frameworks, they do not explicitly consider RL-based optimization over CTMC-based processes, and RL-based optimization for joint continuous-discrete frameworks remains largely underexplored.

\textbf{Multi-Objective Strategies for Molecular Design.}
In multi-objective molecular design, a common approach is to combine multiple objectives into a single scalar reward using Weighted Sum (WS)~\cite{WS}, which has been widely adopted in prior studies such as MolDQN \cite{MolDQN}, MORLD \cite{MORLD}, and DA-MolDQN \cite{DA-MolDQN}. However, WS cannot prevent objective dominance, as the overall update is a linear combination of gradients, where objectives with larger gradient magnitudes tend to dominate the updates, even with carefully chosen weights. Beyond WS-based approaches, some methods address multi-objective conflicts at the gradient level. SMORE-DRL \cite{SMORE-DRL} adopts a Singular Value Decomposition (SVD)\cite{SVD}-based gradient decomposition to capture shared directions across objectives, while PCGrad \cite{PCGrad} projects gradients to remove conflicting components before aggregation. Although these gradient alignment methods mitigate conflicts, dominance may still arise when gradients are non-conflicting but exhibit significant magnitude imbalance.

\section{Methodology}

\begin{figure*}[htbp]
    \centering
    \vspace{-3mm}
    \includegraphics[width=\linewidth]{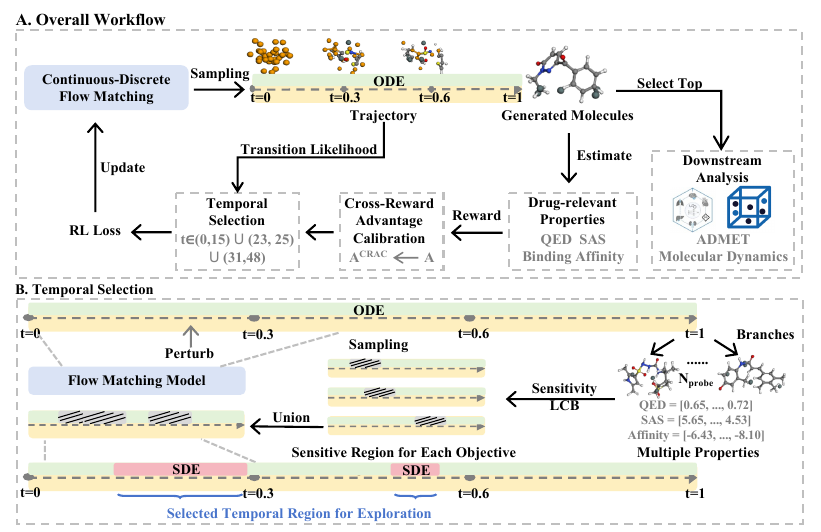}
    \caption{Overview of the proposed RL-guided continuous-discrete flow matching framework with temporal selective exploration. (A) Overall workflow. A continuous-discrete flow matching model is optimized via RL to generate molecules with improved drug-relevant properties. Generated molecules are evaluated using QED, SAS, and binding affinity. CRAC balances the objectives using their remaining gaps to the target criteria and constructs the advantage for model updates. Top generated molecules are further evaluated using ADMET analysis and MD simulation. (B) Temporal selection. Probe trajectories are sampled along the generation process. At each probed timestep, each probe is expanded into multiple perturbed branches to estimate how molecular properties respond to perturbations. The label Perturb denotes generating multiple stochastic branches from the same intermediate molecular state. The ODE and SDE labels describe continuous-coordinate dynamics, while discrete molecular attributes evolve through CTMC transitions. For each objective, sensitivity is estimated using an LCB criterion, and the selected regions are combined into the final temporal window. Stochastic exploration is then applied within the selected window during RL training.}
\vspace{-3mm}
    \label{fig:Architecture}
\end{figure*}

This section presents the core components of our framework, including continuous-discrete flow matching model, temporal selection, CRAC, and the full framework. The overall architecture of our framework is illustrated in Fig.~\ref{fig:Architecture}.

\subsection{Continuous-Discrete Flow Matching}

This section explains the essential components of the continuous-discrete flow matching framework~\cite{FlowMol3}. The full formulation and details are provided in Appendix~\ref{appendix:Continuous_and_Discrete_Flow_Matching}. Specifically, we consider molecular generation as a joint process over continuous atomic coordinates and discrete molecular identities. Let \(g = (X, A, C, E)\), where \(X \in \mathbb{R}^{N \times 3}\) denotes the 3D atomic coordinates, while \(A\), \(C\), and \(E\) denote atom types, formal charges, and bond types, respectively.

\subsubsection{Continuous Flow Matching}

For atomic coordinates, flow matching defines a transport process between an initial state \(X_0\) and a target structure \(X_1\). The intermediate state at timestep \(t \in [0,1]\) is defined as \(X_t = (1-t)X_0 + tX_1\). The corresponding vector field is parameterized as 
\(u_\theta(G_t,t)=\frac{\widehat X_{1,\theta}(G_t,t)-X_t}{1-t}\), 
where \(\widehat X_{1,\theta}(G_t,t)\) denotes the predicted target endpoint.

To enable stochastic exploration during RL optimization, the coordinate transition is
\begin{equation}
X_{t+\Delta t}=X_t+u_\theta(G_t,t)\Delta t+\tfrac12 V_t\widehat{s}_\theta(G_t,t)+\sqrt{V_t}\epsilon,\quad \epsilon\sim\mathcal N(0,I),
\label{eq:sde}
\end{equation}
where $\widehat{s}_\theta(G_t,t)=-(X_t-\alpha_t\widehat X_{1,\theta}(G_t,t))/(1-\alpha_t)^2$. Here, $V_t$ is the variance of the stochastic coordinate transition, and $\widehat{s}_\theta$ is an endpoint-based estimate of the score function. The score-based correction adjusts the drift when stochasticity is introduced into the flow trajectory.

\subsubsection{Discrete Flow Matching}

For discrete molecular attributes, including atom types, charges, and bond types, the generation process is modeled using a CTMC-based discrete flow formulation introduced by Campbell et al.~\cite{campbell2024discrete}. Given the current state \(a_i^{(t)}\), the transition probability over a small timestep \(\Delta t\) is defined as 
\(p(a_i^{(t+\Delta t)} = j \mid A_t)=\delta(j,a_i^{(t)})+u_i(j,A_t)\Delta t\), 
where \(u_i(j,A_t)\) denotes the transition probability velocity toward state \(j\). This small-step transition is used for discrete sampling. The transition dynamics are governed by stochastic masking and unmasking operations, while the target state distribution is parameterized by the predicted endpoint distribution \(p_\theta(a_i^{(1)} \mid A_t)\).

\subsection{Transition Likelihood for Flow Matching Optimization}

To enable RL optimization on continuous-discrete flow matching, the generation process must define valid transition likelihoods over both continuous and discrete transitions. In this section, we derive the corresponding transition likelihood formulations.

\subsubsection{Continuous Transition Likelihood}

Based on Eq.~\ref{eq:sde}, the mean is $\mu_t=X_t+u_\theta(G_t,t)\Delta t+\tfrac12 V_t\widehat{s}_\theta(G_t,t)$. Thus $X_{t+\Delta t}\mid G_t\sim\mathcal N(\mu_t,V_t I)$ and $\log\pi_\theta^{\mathrm{cont}}=\log\mathcal N(X_{t+\Delta t};\mu_t,V_t I)$.

\subsubsection{Discrete CTMC Transition Likelihood}

The discrete transition likelihood is defined over the masking and unmasking process used during generation. At timestep \(t\), the model predicts an endpoint categorical distribution \(p_\theta(z^{(1)} \mid G_t,t)\), where \(G_t\) denotes the current molecular state and \(z\) represents the corresponding discrete molecular attribute. Let \(m\) denote the mask token and \(\Delta t=s-t\) the integration interval. The masking and unmasking transition probabilities are defined as 
\(q_{\mathrm{unmask}}(t)=\Delta t \frac{\alpha_t' + \eta \alpha_t}{1-\alpha_t}\) 
and 
\(q_{\mathrm{mask}}(t)=\Delta t \eta\), 
where \(\alpha_t\) is the interpolant schedule and \(\eta\) controls the stochasticity of the discrete process. Both masking and unmasking probabilities are clipped to \([0,1]\) to ensure valid Bernoulli transitions.

For masked states, the transition first determines whether the token should be unmasked, followed by sampling a target token \(\hat z_1\) from the predicted endpoint distribution. The resulting likelihood is given by \(\log \pi_\theta = \log \mathrm{Bernoulli}(u_t;q_{\mathrm{unmask}}) + \mathbb{I}[u_t=1]\log p_\theta(\hat z_1 \mid G_t,t)\), where \(u_t\) denotes the unmasking decision. For non-masked states, the likelihood only evaluates whether the current token is remasked, i.e., \(\log \pi_\theta = \log \mathrm{Bernoulli}(v_t;q_{\mathrm{mask}})\), where \(v_t\) denotes the masking decision.

The overall discrete transition likelihood is obtained by aggregating the item-level likelihoods over all discrete molecular components, 
\(\log \pi_\theta^{\mathrm{disc}}=\sum_{i\in\mathcal{I}}\log \pi_{\theta,i}\), 
where \(\mathcal{I}\) denotes the set of discrete molecular elements. More details are provided in Appendix~\ref{appendix:Discrete_CTMC_Transition_Derivation}.

\subsubsection{Continuous-Discrete Joint Transition Likelihood}

The overall transition likelihood combines the continuous and discrete components, defined as
\begin{equation}
\log \pi_\theta
=
\log \pi_\theta^{\mathrm{cont}}
+
\log \pi_\theta^{\mathrm{disc}}.
\label{eq:joint_transition}
\end{equation}
This unified formulation enables RL optimization for continuous-discrete flow matching.

\subsection{Temporal Selection for RL Exploration}

Applying stochastic exploration across all timesteps may introduce unnecessary noise, since different timesteps do not contribute equally to the final molecular properties. To improve optimization efficiency, temporal selection restricts stochastic exploration to reward-sensitive temporal regions.

Let the generation process be discretized into \(T\) integration steps with normalized times \(t_0,\ldots,t_{T-1} \in [0,1]\). A probe batch of \(N_{\text{probe}}\) latent samples is first drawn from the flow prior. To reduce computational cost, probe times are uniformly subsampled using a stride parameter, producing a probe set \(\mathcal{T}_{\text{probe}} \subseteq \{t_0,\ldots,t_{T-1}\}\).

For each probe time \(t \in \mathcal{T}_{\text{probe}}\), the trajectory is integrated up to timestep \(t\), producing an intermediate molecular state \(G_t^{(i)}\) for probe sample \(i\). Multiple perturbed rollouts are then independently generated from \(G_t^{(i)}\) by applying stochastic continuous-discrete transitions, producing \(B\) local branches indexed by \(b\). Gaussian perturbation is applied to the continuous coordinates, while stochastic masking and unmasking transitions are sampled for the discrete molecular attributes through the CTMC process. Each perturbed branch is then rolled out to the final state under the generation process, producing multiple endpoint rollouts for every probe sample and timestep.

For objective \(k\), the sensitivity at timestep \(t\) is estimated using the reward variation across perturbed rollouts, defined as \(\Delta_{i,k}(t)=\max_b r_{i,t}^{(b,k)}-\min_b r_{i,t}^{(b,k)}\), where \(r_{i,t}^{(b,k)}\) denotes the reward of branch \(b\) for probe sample \(i\). This range measures how strongly perturbations can affect the final reward, without indicating whether they improve or worsen it. During PPO optimization, sampled transitions associated with positive advantages are reinforced, while those associated with negative advantages are discouraged. To reduce estimation noise, sensitivities are aggregated across probe samples to obtain the empirical mean \(\mu_k(t)\) and standard error \(\mathrm{SE}_k(t)\). A confidence-adjusted sensitivity score is then computed using a lower confidence bound (LCB), 
\(\mathrm{LCB}_k(t)=\mu_k(t)-1.96\,\mathrm{SE}_k(t)\). The reward-sensitive set \(\mathcal{T}_k\) is defined as the timesteps with positive LCB scores. The union across all objectives,
\(\mathcal{W}=\bigcup_{k=1}^{K}\mathcal{T}_k\),
defines the final temporal exploration window. During RL optimization, stochastic exploration is applied only within \(\mathcal{W}\), while all remaining timesteps follow the original generation process. The algorithm and additional analysis are provided in Appendix~\ref{appendix:Adaptive_Stochasticity_analysis}.

\subsection{CRAC}

In multi-objective optimization, a molecule is desirable when it satisfies all target property criteria. CRAC constructs an advantage that accounts for both the remaining gap to each objective and joint satisfaction of the criteria. Let $r_{i,k}\in[0,1]$ be the cutoff-based reward for objective $k$ and molecule $i$, where $r_{i,k}=1$ indicates that the objective has reached its target. We define $v_{i,k}=1-r_{i,k}$ and its batch mean $\bar v_k$. The per-objective advantage is
\begin{equation}
a_{i,k}=\operatorname{clip}\!\left(\frac{\bar v_k-v_{i,k}}{\max(\operatorname{std}_i(v_{i,k}),\epsilon_v)},-c,c\right).
\end{equation}
CRAC assigns greater weight to objectives with larger average violations. The effective weight is
\begin{equation}
\widetilde w_k=\frac{w_k\sqrt{\bar v_k+\epsilon_w}}{\sum_j w_j\sqrt{\bar v_j+\epsilon_w}}.
\end{equation}
Here, $\epsilon_v$ and $\epsilon_w$ are stability constants, and $c$ bounds the per-objective advantage. The weighted per-objective advantages are combined with a joint-success signal, producing $A_i^{\mathrm{CRAC}}$ for the PPO update. The full formulation is provided in Appendix~\ref{appendix:crac}.

\subsection{Full Workflow}
\label{sec:full-rl}

This section presents the full RL-guided continuous-discrete flow matching framework with temporal selective exploration for multi-objective molecular generation. Starting from a pretrained flow matching model \(\theta_{\mathrm{old}}\), we sample a batch of \(N\) molecules and record the corresponding generation trajectories. The generation process is viewed as a Markov chain over molecular states \(G_t\), where each transition is evaluated using the continuous-discrete transition likelihood defined above.

\textbf{Reward design.}
After generation, we evaluate the final molecule \(G_1^{(i)}\) for \(K\) molecular properties. Since molecular properties may exhibit different optimization directions and numerical scales, all property values are first normalized, as detailed in Appendix~\ref{appendix:Ranges_Normalization_and_Scaling}. Our objective is to generate more molecules satisfying all predefined property cutoffs (see Appendix~\ref{appendix:Performance_Metrics}). Accordingly, the reward is defined based on the relative distance to the target cutoff. Let \(p_{i,k}\in[0,1]\) denote the normalized value of objective \(k\) for sample \(i\), and let \(\tau_k\in[0,1]\) denote the corresponding normalized strict cutoff. The reward for each objective is defined as 
\(r_{i,k}=\mathrm{clip}(p_{i,k}-\tau_k+1,\,0,\,1)\).
This design gives full reward if the target cutoff is reached and avoids over-optimization of already satisfactory properties. As a result, the RL update can focus more on objectives that remain below desired thresholds. The resulting reward vector \(\mathbf r_i=(r_{i,1},\ldots,r_{i,K})\) is used to compute \(A_i^{\mathrm{CRAC}}\) as described above.

\textbf{Temporal selective RL update.}
Because the generation process is a Markov chain, the final molecule depends on the sequence of intermediate transitions that produced it. Thus, the terminal advantage \(A_i^{\mathrm{CRAC}}\) can be treated as a delayed return and assigned to the selected transitions along the same trajectory~\cite{RL4generative-DDPO}. Then, for each selected timestep \(t\in\mathcal S\), the likelihood ratio is computed as 
\(\rho_{i,t}=\exp\left(\log \pi_\theta(G_{t+\Delta t}^{(i)}\mid G_t^{(i)})-\log \pi_{\theta_{\mathrm{old}}}(G_{t+\Delta t}^{(i)}\mid G_t^{(i)})\right)\). We adopt a Proximal Policy Optimization (PPO)-style~\cite{PPO} clipped surrogate objective to update the model. The loss is defined as
\begin{equation}
\mathcal{L}_{\mathrm{RL}}
=
-
\frac{1}{N|\mathcal S|}
\sum_{i=1}^{N}
\sum_{t\in\mathcal S}
\min
\left(
\rho_{i,t} A_i^{\mathrm{CRAC}},
\,
\mathrm{clip}(\rho_{i,t},1-\epsilon,1+\epsilon)
A_i^{\mathrm{CRAC}}
\right).
\vspace{-3mm}
\label{eq:rl_loss}
\end{equation}
Here, $\mathcal S$ denotes the timesteps selected for the PPO update. Temporal selection is used during RL training. At inference, the optimized model generates molecules through the full generation process.

\section{Experiments}

\subsection{Experimental Settings}
In this section, we describe the task setup, baselines, and ablation study design. Details of datasets (e.g., sources, descriptions, preprocessing, splitting, statistics, etc.), hyperparameter selection, performance metrics, downstream analysis (ADMET analysis and MD simulation) setups, as well as computational time and resources, are provided in Appendix~\ref{appendix:Additional_Experimental_Details}.

\subsubsection{Task setup}

\textbf{Drug discovery.} 
The primary task is to design small molecules with desired drug-related properties. Specifically, we use RL to optimize a flow matching model pretrained on the GEOM-Drugs dataset, such that the model can generate molecules with improved drug-related properties and identify promising candidates. In this task, we consider three commonly used properties in inhibitor design, namely QED, SAS, and binding affinity. QED and SAS are computed using RDKit~\cite{rdkit2006}, which is a standard approach. Binding affinity is estimated via molecular docking using AutoDock Vina~\cite{autodockvina}, a widely adopted tool in the field, by docking generated molecules to the target proteins. Detailed docking protocols and redocking validation are provided in Appendix~\ref{appendix:Docking_Protocol}. We consider the design of inhibitors for two target proteins, EGFR (PDB ID: 3W2S) and Pin1 (PDB ID: 3IKD), which exhibit different levels of difficulty. Details of these proteins are provided in Appendix~\ref{a ppendix:target_proteins}. Generalization experiment settings are provided in Appendix~\ref{appendix:Generalization_Experiment}.

\subsubsection{Baselines}

We compare our method with baselines from complementary perspectives: the pretrained base model; sampling-based Best-of-$N$ and SMC; RL-guided diffusion (DM+RL), protein-conditioned generation (BInD), and reward-weighted flow matching (CMG); alternative temporal strategies (Random, Uniform, Full-SDE, and Excluded); and multi-objective optimization methods (WS and PCGrad). All methods are evaluated using the same molecular-property criteria. Implementation details are provided in Appendix~\ref{appendix:Baseline_Implementation_and_Analysis}.

\subsubsection{Ablation Study}
\textbf{Component ablation.} To analyze the contribution of each component, we conduct ablation studies from several perspectives under the same training and evaluation setting. Specifically, we consider: (1) w/o discrete: optimizing only the continuous component; (2) w/o continuous: optimizing only the discrete component; (3) w/o selection: optimizing over all timesteps without temporal selection; (4) w/o CRAC: removing CRAC; and (5) w/o both: removing both temporal selection and CRAC.

\textbf{Timestep selection analysis.} 
To further analyze the robustness of the design choices in timestep selection, we conduct additional experiments by varying (1) the set of target properties used for selection, (2) the number of branches, (3) the number of probes, and (4) the random seeds.

\subsection{Results and Discussion}

This section presents the main results, key ablation studies, and downstream analyses. Additional details and results are provided in Appendix~\ref{appendix:Additional_Experiments_Results_and_Discussions}.

\subsubsection{Main Results}
As shown in Table~\ref{tab:main_results}, our method achieves the highest Loose and Strict success rates on both targets while maintaining high molecular quality. On EGFR, it achieves Loose and Strict success rates of 85.70\% and 70.77\%, compared with 78.83\% and 63.50\% for SMC. On Pin1, the corresponding rates are 7.63\% and 0.83\%, compared with 5.83\% and 0.57\% for SMC. SMC also exhibits lower uniqueness on both targets. These results show that our framework improves the joint optimization of molecular properties across targets with different levels of design difficulty.

\definecolor{sectiongray}{RGB}{220,220,220}
\definecolor{bestpink}{RGB}{229,209,170}
\definecolor{secondpink}{RGB}{249,239,216}
\providecommand{\best}[1]{\cellcolor{bestpink}#1}
\providecommand{\second}[1]{\cellcolor{secondpink}#1}
\providecommand{\std}[1]{{\footnotesize $\pm$ #1}}

\begin{table*}[htbp]
\centering
\small
\setlength{\tabcolsep}{3pt}
\vspace{-2mm}
\caption{Performance of baselines and our method. \colorbox{bestpink}{Best} and \colorbox{secondpink}{second-best} results are highlighted.}
\label{tab:main_results}
\resizebox{\textwidth}{!}{%
\begin{tabular}{lcccccccc}
\toprule
\textbf{Method} & \textbf{Validity} & \textbf{Uniqueness} & \textbf{Novelty} & \textbf{VUN} & \textbf{Atom stab.} & \textbf{Mol. stab.} & \textbf{Loose} & \textbf{Strict} \\
\midrule
\rowcolor{sectiongray}
\multicolumn{9}{c}{\textbf{3W2S}} \\
Base Model & 99.37 \std{0.15} & \second{99.97 \std{0.06}} & \best{100.00 \std{0.00}} & 99.33 \std{0.12} & \second{99.99 \std{0.00}} & 99.77 \std{0.06} & 70.50 \std{0.85} & 47.53 \std{0.99} \\
Best-of-N & \best{100.00 \std{0.00}} & 98.80 \std{0.10} & \best{100.00 \std{0.00}} & 98.80 \std{0.10} & \best{100.00 \std{0.00}} & \second{99.87 \std{0.06}} & 74.33 \std{1.67} & 62.33 \std{0.29} \\
SMC & \best{100.00 \std{0.00}} & 65.33 \std{0.47} & \best{100.00 \std{0.00}} & 65.33 \std{0.47} & \second{99.99 \std{0.00}} & 99.63 \std{0.06} & 78.83 \std{2.30} & 63.50 \std{1.51} \\
DM+RL & 94.17 \std{1.72} & 99.08 \std{0.40} & \best{100.00 \std{0.00}} & 93.30 \std{1.75} & 99.54 \std{0.13} & 98.37 \std{0.42} & 72.17 \std{1.72} & 58.47 \std{0.76} \\
BInD & 84.37 \std{0.25} & 98.78 \std{0.13} & \second{99.00 \std{0.07}} & 82.50 \std{0.17} & 96.92 \std{0.18} & 98.77 \std{0.09} & 56.77 \std{0.40} & 39.27 \std{0.38} \\
CMG & 99.03 \std{0.29} & 99.93 \std{0.06} & \best{100.00 \std{0.00}} & 98.97 \std{0.32} & 99.98 \std{0.01} & 99.40 \std{0.17} & 60.03 \std{0.35} & 38.93 \std{0.65} \\
Random & 98.53 \std{0.40} & \best{100.00 \std{0.00}} & \best{100.00 \std{0.00}} & 98.53 \std{0.40} & 99.98 \std{0.00} & 99.40 \std{0.17} & 68.00 \std{1.51} & 44.90 \std{0.70} \\
Uniform & 98.67 \std{0.40} & \best{100.00 \std{0.00}} & \best{100.00 \std{0.00}} & 98.67 \std{0.40} & 99.97 \std{0.01} & 99.23 \std{0.35} & 67.63 \std{2.62} & 44.40 \std{2.29} \\
Full-SDE & 99.37 \std{0.45} & \best{100.00 \std{0.00}} & \best{100.00 \std{0.00}} & \second{99.37 \std{0.45}} & \best{100.00 \std{0.00}} & \best{99.90 \std{0.10}} & 72.33 \std{1.90} & 49.87 \std{1.66} \\
WS & 99.43 \std{0.06} & 99.83 \std{0.15} & \best{100.00 \std{0.00}} & 99.27 \std{0.12} & \best{100.00 \std{0.00}} & 99.87 \std{0.06} & \second{83.27 \std{0.64}} & 63.33 \std{1.50} \\
PCGrad & 99.43 \std{0.06} & 99.63 \std{0.12} & \best{100.00 \std{0.00}} & 99.07 \std{0.06} & \best{100.00 \std{0.00}} & \best{99.90 \std{0.00}} & 82.83 \std{1.03} & \second{67.37 \std{1.11}} \\
Excluded & 98.37 \std{0.38} & \best{100.00 \std{0.00}} & \best{100.00 \std{0.00}} & 98.37 \std{0.38} & 99.98 \std{0.01} & 99.47 \std{0.25} & 66.20 \std{0.95} & 42.93 \std{1.29} \\
\textbf{Ours} & \second{99.77 \std{0.06}} & 99.67 \std{0.12} & \best{100.00 \std{0.00}} & \best{99.43 \std{0.15}} & \best{100.00 \std{0.00}} & \best{99.90 \std{0.10}} & \best{85.70 \std{0.66}} & \best{70.77 \std{0.21}} \\
\midrule
\rowcolor{sectiongray}
\multicolumn{9}{c}{\textbf{3IKD}} \\
Base Model & 99.37 \std{0.15} & \second{99.97 \std{0.06}} & \best{100.00 \std{0.00}} & 99.33 \std{0.12} & \second{99.99 \std{0.00}} & 99.77 \std{0.06} & 4.97 \std{0.64} & 0.37 \std{0.12} \\
Best-of-N & \best{100.00 \std{0.00}} & 97.77 \std{0.12} & \best{100.00 \std{0.00}} & 97.77 \std{0.12} & \best{100.00 \std{0.00}} & \best{100.00 \std{0.00}} & 5.77 \std{0.40} & 0.57 \std{0.06} \\
SMC & \best{100.00 \std{0.00}} & 68.77 \std{0.40} & \best{100.00 \std{0.00}} & 68.77 \std{0.40} & \second{99.99 \std{0.00}} & 99.47 \std{0.06} & 5.83 \std{0.35} & \second{0.57 \std{0.06}} \\
DM+RL & 94.13 \std{1.85} & 99.12 \std{0.33} & \best{100.00 \std{0.00}} & 93.30 \std{1.78} & 99.41 \std{0.07} & 98.46 \std{0.35} & 5.40 \std{0.82} & 0.50 \std{0.10} \\
BInD & 76.20 \std{0.69} & 94.19 \std{0.67} & \second{98.37 \std{0.43}} & 70.60 \std{0.62} & 94.93 \std{0.29} & 91.50 \std{0.27} & 2.03 \std{0.21} & 0.13 \std{0.06} \\
CMG & 97.10 \std{0.40} & 99.90 \std{0.10} & \best{100.00 \std{0.00}} & 97.00 \std{0.36} & 99.98 \std{0.01} & 99.40 \std{0.17} & 5.23 \std{0.06} & 0.47 \std{0.06} \\
Random & 98.20 \std{0.26} & \best{100.00 \std{0.00}} & \best{100.00 \std{0.00}} & 98.20 \std{0.26} & 99.98 \std{0.00} & 99.33 \std{0.12} & 3.27 \std{0.55} & 0.23 \std{0.06} \\
Uniform & 98.20 \std{0.60} & \best{100.00 \std{0.00}} & \best{100.00 \std{0.00}} & 98.20 \std{0.60} & \second{99.99 \std{0.01}} & 99.50 \std{0.36} & 5.07 \std{0.25} & 0.50 \std{0.20} \\
Full-SDE & 98.80 \std{0.00} & \best{100.00 \std{0.00}} & \best{100.00 \std{0.00}} & 98.80 \std{0.00} & \second{99.99 \std{0.00}} & 99.50 \std{0.10} & \second{6.27 \std{0.45}} & 0.40 \std{0.17} \\
WS & \second{99.70 \std{0.10}} & 99.77 \std{0.12} & \best{100.00 \std{0.00}} & \second{99.47 \std{0.15}} & \best{100.00 \std{0.01}} & \second{99.90 \std{0.17}} & 4.47 \std{0.31} & 0.33 \std{0.06} \\
PCGrad & 99.40 \std{0.26} & 99.93 \std{0.06} & \best{100.00 \std{0.00}} & 99.33 \std{0.25} & \second{99.99 \std{0.00}} & 99.83 \std{0.06} & 5.67 \std{0.06} & 0.50 \std{0.00} \\
Excluded & 98.50 \std{0.80} & \best{100.00 \std{0.00}} & \best{100.00 \std{0.00}} & 98.50 \std{0.80} & 99.98 \std{0.01} & 99.10 \std{0.61} & 4.97 \std{0.06} & 0.37 \std{0.06} \\
\textbf{Ours} & 99.63 \std{0.15} & \second{99.97 \std{0.06}} & \best{100.00 \std{0.00}} & \best{99.60 \std{0.17}} & \second{99.99 \std{0.01}} & 99.53 \std{0.29} & \best{7.63 \std{0.42}} & \best{0.83 \std{0.06}} \\
\bottomrule
\end{tabular}}
\vspace{0.5pt}
\raggedright
\footnotesize
\textbf{Note.} Results are reported as mean $\pm$ standard deviation over three independent runs with different random seeds, with 1,000 generated molecules per run. Loose and Strict success rates measure the proportions of molecules satisfying the corresponding joint QED, SAS, and binding-affinity criteria. Loose: QED $\geq 0.5$, SAS $\leq 6$, and affinity $\leq -7.0$ kcal/mol. Strict: QED $\geq 0.6$, SAS $\leq 4$, and affinity $\leq -7.5$ kcal/mol. DM+RL denotes the RL-guided diffusion baseline. Detailed thresholds and evaluation procedures are provided in Appendix~\ref{appendix:Performance_Metrics}.
\end{table*}

\subsubsection{Ablation Study}
As shown in Table~\ref{tab:ablation_results}, removing temporal selection or CRAC reduces the Loose and Strict success rates on both targets. On EGFR, the Strict success rate decreases from 70.77\% to 49.87\% without temporal selection and to 56.43\% without CRAC. On Pin1, it decreases from 0.83\% to 0.40\% and 0.43\%, respectively. These results support the contribution of both components to multi-objective optimization.

\begin{figure*}[htbp]
    \centering
    \vspace{-4mm}
    \includegraphics[width=\linewidth]{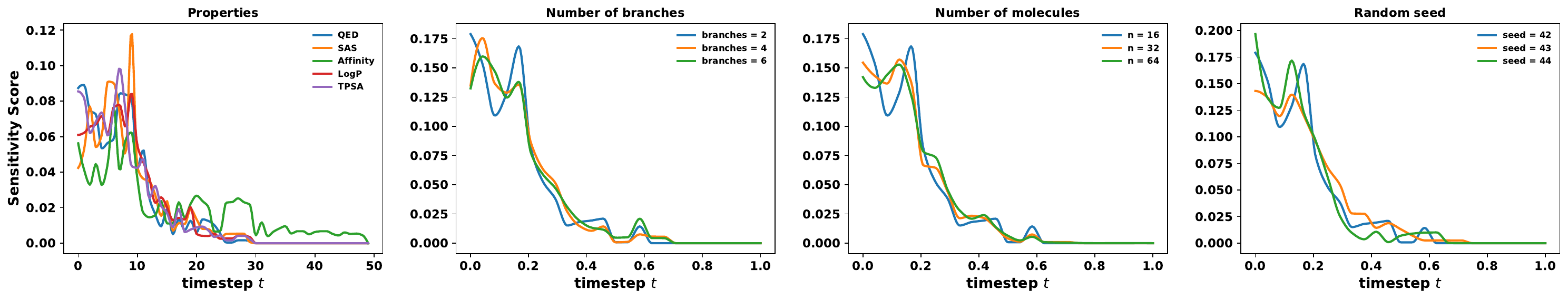}
    \caption{Temporal selection ablation. The analyses of branches, number of probes, and seeds use QED as the target property.}
\vspace{-1mm}
    \label{fig:ablation_property}
\end{figure*}

Figure~\ref{fig:ablation_property} shows that different molecular properties exhibit non-uniform temporal sensitivity along the flow trajectory. Different properties are affected by different temporal regions. In addition, similar sensitive regions are consistently observed under different numbers of branches, probe molecules, and random seeds, indicating the temporal selection process is stable and robust.

\definecolor{sectiongray}{RGB}{220,220,220}
\definecolor{bestpink}{RGB}{229,209,170}
\definecolor{secondpink}{RGB}{249,239,216}
\providecommand{\best}[1]{\cellcolor{bestpink}#1}
\providecommand{\second}[1]{\cellcolor{secondpink}#1}
\providecommand{\std}[1]{{\footnotesize $\pm$ #1}}
\begin{table*}[htbp]
\centering
\small
\setlength{\tabcolsep}{3pt}
\vspace{-2mm}
\caption{Ablation study. \colorbox{bestpink}{\strut Best} and \colorbox{secondpink}{\strut second-best} results are highlighted.}
\vspace{-3mm}
\label{tab:ablation_results}
\resizebox{\textwidth}{!}{%
\begin{tabular}{lcccccccc}
\toprule
\textbf{Method} & \textbf{Validity} & \textbf{Uniqueness} & \textbf{Novelty} & \textbf{VUN} & \textbf{Atom stab.} & \textbf{Mol. stab.} & \textbf{Loose} & \textbf{Strict} \\
\midrule
\rowcolor{sectiongray}
\multicolumn{9}{c}{\textbf{3W2S}} \\
w/o discrete & 98.77 \std{0.21} & 99.80 \std{0.10} & \best{100.00 \std{0.00}} & 98.57 \std{0.15} & 99.58 \std{0.07} & 99.49 \std{0.11} & \second{79.37 \std{1.15}} & \second{58.57 \std{1.63}} \\
w/o continuous & 98.33 \std{0.15} & 99.29 \std{0.10} & \best{100.00 \std{0.00}} & 97.63 \std{0.12} & 99.96 \std{0.01} & \second{99.88 \std{0.06}} & 77.07 \std{1.25} & 57.93 \std{1.80} \\
w/o temporal selection & 99.37 \std{0.45} & \best{100.00 \std{0.00}} & \best{100.00 \std{0.00}} & 99.37 \std{0.45} & \best{100.00 \std{0.00}} & \best{99.90 \std{0.10}} & 72.33 \std{1.90} & 49.87 \std{1.66} \\
w/o CRAC & \second{99.60 \std{0.26}} & \best{100.00 \std{0.00}} & \best{100.00 \std{0.00}} & \best{99.60 \std{0.26}} & \second{99.99 \std{0.01}} & 99.63 \std{0.25} & 78.13 \std{1.32} & 56.43 \std{1.61} \\
w/o both & 98.23 \std{0.32} & \second{100.00 \std{0.00}} & \best{100.00 \std{0.00}} & 98.23 \std{0.32} & 99.97 \std{0.02} & 99.42 \std{0.26} & 71.07 \std{1.26} & 45.60 \std{1.71} \\
\textbf{Ours} & \best{99.77 \std{0.06}} & 99.67 \std{0.12} & \best{100.00 \std{0.00}} & \second{99.43 \std{0.15}} & \best{100.00 \std{0.00}} & \best{99.90 \std{0.10}} & \best{85.70 \std{0.66}} & \best{70.77 \std{0.21}} \\
\midrule
\rowcolor{sectiongray}
\multicolumn{9}{c}{\textbf{3IKD}} \\
w/o discrete & 98.10 \std{0.17} & 99.73 \std{0.12} & \best{100.00 \std{0.00}} & 97.83 \std{0.15} & \best{99.99 \std{0.01}} & \second{99.57 \std{0.13}} & 5.90 \std{0.10} & 0.40 \std{0.10} \\
w/o continuous & 98.80 \std{0.26} & 99.80 \std{0.10} & \best{100.00 \std{0.00}} & 98.60 \std{0.17} & \best{99.99 \std{0.00}} & 99.46 \std{0.09} & 5.87 \std{0.35} & \second{0.43 \std{0.15}} \\
w/o temporal selection & \second{98.80 \std{0.00}} & \best{100.00 \std{0.00}} & \best{100.00 \std{0.00}} & \second{98.80 \std{0.00}} & \best{99.99 \std{0.00}} & 99.50 \std{0.10} & \second{6.27 \std{0.45}} & 0.40 \std{0.17} \\
w/o CRAC & \best{99.63 \std{0.21}} & \second{99.97 \std{0.06}} & \best{100.00 \std{0.00}} & \best{99.60 \std{0.26}} & \best{99.99 \std{0.00}} & 99.50 \std{0.17} & 5.90 \std{0.20} & \second{0.43 \std{0.12}} \\
w/o both & 98.20 \std{0.30} & \best{100.00 \std{0.00}} & \best{100.00 \std{0.00}} & 98.20 \std{0.30} & \best{99.99 \std{0.00}} & \best{99.75 \std{0.13}} & 5.03 \std{0.06} & 0.43 \std{0.06} \\
\textbf{Ours} & \best{99.63 \std{0.15}} & \second{99.97 \std{0.06}} & \best{100.00 \std{0.00}} & \best{99.60 \std{0.17}} & \best{99.99 \std{0.01}} & 99.53 \std{0.29} & \best{7.63 \std{0.42}} & \best{0.83 \std{0.06}} \\
\bottomrule
\end{tabular}}
\vspace{0.5pt}
\raggedright
\footnotesize
\textbf{Note.} Results are reported as mean $\pm$ standard deviation over three independent runs with different random seeds, with 1,000 generated molecules per run.
\end{table*}

\subsubsection{Docking, ADMET Analysis and MD simulations}

\begin{figure*}[htbp]
    \centering
    \includegraphics[width=0.9\linewidth]{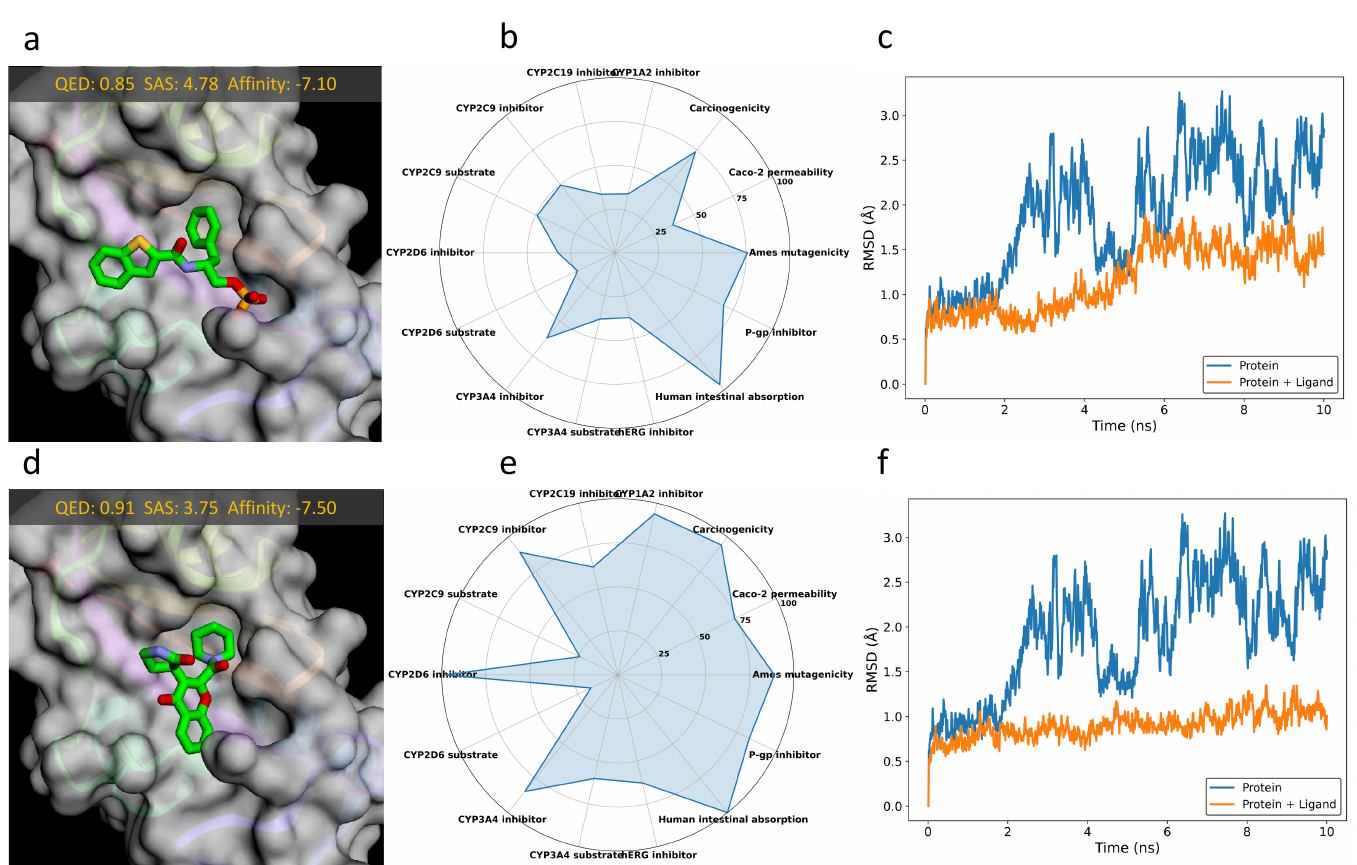}
    \caption{Comparison between the reference ligand and the generated molecule for the Pin1 (3IKD) system. The three subfigures in each row represent docking visualization, ADMET analysis (farther from center is better), and MD simulation RMSD trajectories (lower is more stable) from left to right, respectively. (a--c) Co-crystallized reference ligand. (d--f) The top generated molecule.}
\label{fig:downstream_main}
\end{figure*}

As shown in Fig.~\ref{fig:downstream_main}, the top generated molecule (selected based on properties) overall exhibits improved docking, ADMET, and MD stability performance compared to the reference ligand in the Pin1 (3IKD) system. Specifically, the generated molecule achieves higher drug-likeness (\(0.91\) vs. \(0.85\)), lower synthetic accessibility score (\(3.75\) vs. \(4.78\)), and stronger binding affinity (\(-7.50\) vs. \(-7.10\)). In the ADMET analysis, the generated molecule shows improved overall profiles across most evaluated ADMET properties, including lower predicted Ames mutagenicity and carcinogenicity risks, together with competitive permeability-related profiles. These results suggest improved drug development potential. In addition, the MD simulations show that the generated complex maintains lower RMSD values and reduced fluctuations throughout the simulation compared to the reference complex, indicating more stable binding within the Pin1 binding pocket. It suggests that this molecule may serve as a promising candidate for further experimental synthesis and validation.

Overall, these observations demonstrate that the proposed framework can generate molecules with stable binding behavior, good drug-related properties, and synthesis-friendly structures, even for challenging targets such as Pin1. More generated results are provided in Appendix~\ref{appendix:Additional_Experiments_Results_and_Discussions}.

\section{Conclusion}
\vspace{-2mm}We presented an RL-guided continuous-discrete flow matching framework for multi-objective 3D molecular generation. Our temporal selection strategy identifies property-sensitive regions of the generation trajectory and focuses stochastic exploration on these regions, while CRAC balances multiple objectives during optimization. Experiments on EGFR and Pin1 show improved joint success rates while maintaining high molecular quality. Additional in silico analyses of generated candidates further demonstrate the applicability of the framework. These results support the potential of the approach for early-stage molecular discovery.

\begin{ack}
This work was supported in part by the Canada Research Chairs Tier II Program (CRC-2021-00482), the Canadian Institutes of Health Research (PJT 190272, PJT 204042, CFA-205059), the Natural Sciences and Engineering Research Council of Canada (RGPIN-2026-07632, ALLRP 602759-24), the New Frontiers in Research Fund--Exploration (NFRFE-2025-01310), and the Canada Foundation for Innovation John R. Evans Leaders Fund program (\#43481).
\end{ack}

\bibliography{references}
\bibliographystyle{unsrtnat}

\appendix

\newpage
\appendix

\counterwithin{figure}{section}
\counterwithin{table}{section}
\counterwithin{equation}{section}
\counterwithin{algorithm}{section}

{
\titleformat{\part}[display]
  {\normalfont\Large\bfseries}
  {}
  {0pt}
  {}

\part*{Appendix}
\addcontentsline{toc}{part}{Appendix}

\etocsetnexttocdepth{2}
\localtableofcontents
}

\clearpage
\section{Design Choices and Rationale}
\label{appendix:Design_Choices_and_Rationale}
\subsection{Design Choice: Unconditional Generation}

Our model focuses on unconditional molecular generation rather than property-conditioned generation. This design allows the model to explore the chemical space more broadly without being restricted by predefined conditions, leading to more diverse candidate molecules and increasing the likelihood of discovering useful structures. As summarized in Table~\ref{tab:conditional_vs_unconditional}, property-conditioned generation is typically designed to match specific target properties and is therefore more constrained. In contrast, unconditional generation provides greater flexibility for large-scale candidate exploration, as it can be readily applied to different property objectives or target proteins without requiring a new pretrained generative backbone, while still allowing task-specific RL fine-tuning.

\begin{table}[htbp]
\centering
\caption{Comparison between conditional and unconditional generation.}
\label{tab:conditional_vs_unconditional}
\begin{tabular}{lcc}
\toprule
\textbf{Aspect} & \textbf{Conditional} & \textbf{Unconditional} \\
\midrule
Condition inputs & Required (property values) & Not required \\
Candidate diversity & Moderate & High \\
Large-scale exploration & Moderate & Strong \\
Adaptation to new tasks & \makecell{May require additional\\conditioning or retraining} & \makecell{Shared backbone for\\task-specific optimization} \\
\bottomrule
\end{tabular}
\end{table}

\subsection{Choice of Optimization Framework}
\label{appendix:Choice_of_Optimization_Framework}
We use a PPO-style objective for this work. GRPO constructs advantages relative to a group of sampled outputs, and its behavior depends on how groups are defined. Our unconditional molecular generation setting does not provide a natural shared prompt or target input for defining such groups, so we adopt PPO-style optimization in our experiments. A summary comparison is provided in Table~\ref{tab:ppo_vs_grpo}.

\begin{table}[htbp]
\centering
\caption{Comparison between PPO and GRPO under our setting.}
\label{tab:ppo_vs_grpo}
\begin{tabular}{lcc}
\toprule
\textbf{Aspect} & \textbf{PPO} & \textbf{GRPO} \\
\midrule
Optimization signal & Advantage-based & Group-relative advantage \\
Grouping requirement & Not required & Required \\
Shared condition & Not required & Often used \\
Unconditional generation & Used in this work & Group definition needed \\
Batch dependence & Depends on estimator & Depends on group \\
\bottomrule
\end{tabular}
\end{table}

\section{Additional Methodology Details}
\subsection{Continuous and Discrete Flow Matching}
\label{appendix:Continuous_and_Discrete_Flow_Matching}

\subsubsection{Overview}
In this section, we provide a detailed formulation of the flow matching backbone used for 3D molecular generation. The goal of this section is to define the generative process over molecular graphs, including the continuous transport of atomic coordinates, the discrete evolution of molecular identities, and the joint training objective that couples these modalities. This formulation serves as the foundation for the molecular generator used in our experiments.

A molecule is naturally a mixed-type object. Its 3D geometry is continuous, while its chemical identity is described by discrete variables such as atom types, formal charges, and bond orders. Therefore, a generative model for 3D molecules must jointly model continuous and categorical components. The backbone considered here follows this multi-modal perspective. It represents each molecule as a molecular graph and defines a time-dependent generative process that transports a simple prior distribution into the data distribution of valid molecular structures.

The key idea of flow matching is to learn a vector field, or its discrete analogue, that moves samples from a source distribution to a target distribution over a normalized time interval $t \in [0,1]$. At $t=0$, samples are drawn from a simple prior. At $t=1$, samples are expected to match molecules from the data distribution. For continuous coordinates, this transport is described by an ODE. For categorical molecular attributes, the evolution is described by CTMCs. These two components are trained together so that the model can generate a complete molecular graph with both realistic geometry and chemically meaningful discrete structure.

In the following subsections, we first introduce the molecular representation and notation. We then describe the continuous flow matching formulation for atomic coordinates, followed by the discrete flow matching formulation for atom types, formal charges, and bond orders. After that, we explain how the continuous and discrete processes are combined into a single multi-modal molecular generation process. Finally, we present the training pipeline, including the model input, model output, and the overall flow matching training objective.

\subsubsection{Notation and Molecular Representation}
\label{appendix:Notation_and_Molecular_Representation}

A molecule is represented as a fully-connected graph with $N$ atoms, where each atom corresponds to a node and each pair of atoms is associated with an edge. The molecular state is defined as a tuple
\begin{equation}
g = (X, A, C, E),
\end{equation}
where each component corresponds to a different modality of the molecule.

The atomic coordinates are denoted by
\begin{equation}
X = \{x_i\}_{i=1}^N \in \mathbb{R}^{N \times 3},
\end{equation}
where $x_i \in \mathbb{R}^3$ represents the 3D position of atom $i$. This is a continuous variable that captures the geometric structure of the molecule.

The discrete molecular attributes consist of atom types, formal charges, and bond orders. Specifically, the atom types are defined as
\begin{equation}
A = \{a_i\}_{i=1}^N, \quad a_i \in \{1, \dots, K_A\},
\end{equation}
where $K_A$ is the number of possible atom types. Similarly, the formal charges are given by
\begin{equation}
C = \{c_i\}_{i=1}^N, \quad c_i \in \{1, \dots, K_C\},
\end{equation}
and the bond orders are defined for each pair of atoms as
\begin{equation}
E = \{e_{ij}\}_{i \neq j}, \quad e_{ij} \in \{1, \dots, K_E\}.
\end{equation}
All discrete variables are represented using one-hot encodings in practice, enabling efficient processing by neural networks.

A time-dependent stochastic process over molecular states is introduced, denoted by $\{g_t\}_{t \in [0,1]}$. At time $t=0$, the process is initialized from a simple prior distribution
\begin{equation}
g_0 \sim p_0(g),
\end{equation}
where the coordinates are sampled from an isotropic Gaussian distribution and all discrete variables are initialized to a special masking state. At time $t=1$, the process aims to match the data distribution of real molecules
\begin{equation}
g_1 \sim p_{\text{data}}(g).
\end{equation}

To enable tractable learning, a coupling distribution is introduced over pairs of initial and target states
\begin{equation}
(g_0, g_1) \sim p(g_0, g_1),
\end{equation}
which specifies how samples from the prior are associated with data samples. Based on these paired states, a family of intermediate distributions $p_t(g \mid g_0, g_1)$ is defined to describe the evolution of molecular states over time.

Throughout this formulation, $t \in [0,1]$ denotes the normalized time variable, and $\Delta t$ denotes a small integration step. The generative process is constructed such that samples evolve smoothly from $g_0$ to $g_1$ as $t$ increases, while maintaining consistency between continuous geometry and discrete molecular structure.

This multi-modal formulation enables molecular generation to be treated as a unified transport problem over both continuous and discrete variables, which will be formalized in the following sections.

\subsubsection{Continuous Flow Matching Formulation for 3D Atomic Coordinates}
\label{app:continuous_flow}

The continuous component of the generative process models the evolution of atomic coordinates $X_t \in \mathbb{R}^{N \times 3}$ over time. The objective is to transport samples from a simple prior distribution $p_0(X)$ to the data distribution $p_{\text{data}}(X)$ by defining a time-dependent process on the interval $t \in [0,1]$.

To construct such a process, pairs of initial and target coordinates $(X_0, X_1)$ are drawn from a coupling distribution. Based on these paired states, an intermediate trajectory is defined using a deterministic interpolation
\begin{equation}
X_t = (1 - t) X_0 + t X_1,
\label{eq:linear_interpolation}
\end{equation}
which describes a straight-line path between $X_0$ and $X_1$ in $\mathbb{R}^{N \times 3}$. This formulation induces a time-indexed family of intermediate states that smoothly connects the prior and data distributions.

The associated vector field is obtained by differentiating the trajectory with respect to time,
\begin{equation}
u_t(X_t; X_0, X_1) = \frac{dX_t}{dt} = X_1 - X_0.
\end{equation}
This expression can be equivalently rewritten in terms of the current state $X_t$ as
\begin{equation}
u_t(X_t; X_0, X_1) = \frac{X_1 - X_t}{1 - t},
\label{eq:vector_field}
\end{equation}
which highlights that the instantaneous velocity depends on both the current state and the target endpoint.

The generative model aims to approximate the marginal vector field $u_t(X_t)$ using a neural network. Instead of directly predicting the velocity, an endpoint parameterization is adopted, where the model predicts the final coordinates $\hat{X}_1(G_t, t)$. The vector field is then defined as
\begin{equation}
u_{\theta}(G_t, t) = \frac{\hat{X}_1(G_t, t) - X_t}{1 - t}.
\label{eq:endpoint_parameterization}
\end{equation}

Training is carried out by minimizing the discrepancy between the predicted endpoint $\hat{X}_1$ and the true endpoint $X_1$ under the interpolated trajectory. This leads to the continuous flow matching objective
\begin{equation}
\mathcal{L}_{X}
=
\mathbb{E}_{t \sim \mathcal{U}(0,1),\; (G_0,G_1),\; G_t}
\left[
w(t)\,
\| \hat{X}_1(G_t, t) - X_1 \|^2
\right],
\label{eq:continuous_loss}
\end{equation}
where $w(t)$ is a time-dependent weighting function that adjusts the contribution of different time steps.

Once trained, the model defines a continuous-time transport process. During generation, the coordinates are updated by numerically integrating the learned vector field
\begin{equation}
X_{t+\Delta t} = X_t + u_{\theta}(G_t, t)\,\Delta t,
\label{eq:ode_update}
\end{equation}
starting from an initial sample $X_0 \sim p_0(X)$.

The update above is deterministic. For stochastic exploration, we use the score-corrected transition in Eq.~\ref{eq:sde}. For $s=t+\Delta t$, its dance-style single-step variance is $V_{t,s}=\eta_x^2(\alpha_s-\alpha_t)$, where $\eta_x$ controls continuous-coordinate noise and is distinct from the discrete CTMC stochasticity parameter. The implementation applies numerical safeguards to the variance and the score denominator near the endpoints.

\subsubsection{Discrete Flow Matching Formulation for Molecular Identity}
\label{app:discrete_flow}

In addition to continuous atomic coordinates, molecular graphs contain discrete attributes, including atom types, formal charges, and bond orders. These variables evolve over time through a stochastic process defined on a discrete state space.

Each discrete modality is represented as a collection of categorical variables. For example, atom types are given by $A_t = \{a_i^{(t)}\}_{i=1}^N$, where each $a_i^{(t)}$ takes values in a finite set of size $K_A$. Similar definitions apply to formal charges $C_t$ and bond orders $E_t$. For notational simplicity, the following formulation is described for a generic discrete variable, with the understanding that it applies independently to each modality.

The evolution of discrete variables is modeled using a CTMC over the interval $t \in [0,1]$. At time $t=0$, all variables are initialized to a special masking state, denoted by $M$. At time $t=1$, the process aims to recover the target discrete configuration.

To construct the trajectory, pairs of initial and target states $(A_0, A_1)$ are sampled from a coupling distribution. Based on these paired states, a time-dependent distribution over intermediate states is defined such that, for each element,
\begin{equation}
p_t(a_i) = (1 - t)\,\delta(a_i = M) + t\,\delta(a_i = a_i^{(1)}),
\label{eq:discrete_path}
\end{equation}
where $\delta(\cdot)$ denotes the Kronecker delta. This formulation describes a gradual transition from the masked state to the target value as $t$ increases.

The CTMC is defined through a probability velocity function that determines the instantaneous transition rates between states. At each time step, the transition probability from the current state $a_i^{(t)}$ to a new state $j$ over a small interval $\Delta t$ is given by
\begin{equation}
p(a_i^{(t+\Delta t)} = j \mid A_t)
=
\delta(j, a_i^{(t)}) + u_i(j, A_t)\,\Delta t,
\label{eq:ctmc_transition}
\end{equation}
where $u_i(j, A_t)$ is the probability velocity toward state $j$.

The probability velocity is parameterized using a neural network that predicts a distribution over the final states. Let $p_{\theta}(a_i^{(1)} \mid A_t)$ denote the predicted distribution over target values given the current state. Using this prediction, the transition dynamics are defined such that masked elements are progressively replaced with sampled values from this distribution, while unmasked elements may revert to the masked state with a small probability. This mechanism ensures stochastic exploration while maintaining a bias toward the target configuration.

Training is performed by maximizing the likelihood of the true discrete values under the predicted endpoint distribution. This leads to a cross-entropy objective of the form
\begin{equation}
\mathcal{L}_{\text{disc}}
=
\mathbb{E}_{t,\; (A_0,A_1),\; A_t}
\left[
- \sum_{i=1}^N
\log p_{\theta}(a_i^{(1)} \mid A_t)\,\mathbf{1}(a_i^{(t)} = M)
\right],
\label{eq:discrete_loss}
\end{equation}
where $\mathbf{1}(\cdot)$ is the indicator function that selects only the masked elements.

During generation, the discrete variables are updated by iteratively sampling from the transition distribution in Eq.~\eqref{eq:ctmc_transition}. At each step, masked elements are replaced with predicted values according to the learned distribution, while unmasked elements may transition back to the masked state depending on the stochastic dynamics. This iterative procedure gradually constructs a complete discrete molecular configuration as $t$ approaches $1$.

This formulation enables the modeling of discrete molecular attributes using a principled stochastic process that is consistent with the flow matching framework, while remaining compatible with the continuous evolution of atomic coordinates.

\subsubsection{Joint Continuous–Discrete Formulation}
\label{appendix: Joint_Continuous–Discrete_Formulation}

The continuous coordinates and discrete molecular attributes together define a unified generative process over molecular graphs. Let $g = (X, A, C, E)$ denote the full molecular state. A time-dependent stochastic process $\{g_t\}_{t \in [0,1]}$ is constructed such that $g_0$ follows a simple prior distribution and $g_1$ matches the data distribution of molecular graphs.

To define the evolution of the full system, pairs of initial and target states $(g_0, g_1)$ are sampled from a coupling distribution. Based on these paired states, a joint trajectory is constructed by combining the continuous interpolation of atomic coordinates with the discrete evolution of molecular attributes.

A key design choice is to factorize the trajectory over modalities. The intermediate distribution over molecular states is defined as
\begin{equation}
p_t(g \mid g_0, g_1)
=
\prod_{m \in \{X, A, C, E\}} p_t(m \mid m_0, m_1),
\label{eq:multimodal_factorization}
\end{equation}
where $m$ denotes one of the modalities and $m_t$ represents its state at time $t$. This formulation assumes that, given the paired endpoints $(g_0, g_1)$, the evolution of each modality can be constructed independently.

Under this factorization, the atomic coordinates follow the continuous trajectory defined in Eq.~\eqref{eq:linear_interpolation}, while the discrete attributes evolve according to the stochastic dynamics described in Eq.~\eqref{eq:ctmc_transition}. Although each modality evolves according to its own dynamics, all modalities share the same time variable and are updated synchronously during generation.

A single neural network is used to model the evolution of all modalities jointly. Given the current state $G_t = g_t$ and time $t$, the network produces a prediction of the final molecular state
\begin{equation}
\hat{G}_1(G_t, t) = \big(\hat{X}_1, \hat{A}_1, \hat{C}_1, \hat{E}_1\big),
\label{eq:joint_prediction}
\end{equation}
where each component corresponds to the predicted endpoint of the respective modality. The prediction for atomic coordinates is used to construct the continuous vector field, while the predictions for discrete attributes define categorical distributions over target states.

This joint parameterization allows information to be shared across modalities. In particular, the prediction of atomic coordinates depends on the discrete structure, and conversely, the prediction of discrete attributes is influenced by the current geometric configuration. As a result, the model captures the strong dependencies between molecular geometry and chemical identity.

During generation, the full molecular state is updated iteratively. At each time step, the continuous coordinates are advanced using the learned vector field, while the discrete attributes are updated by sampling from their respective transition distributions. All modalities are updated at every step, ensuring that the geometry and discrete structure evolve consistently.

\subsubsection{Training Objective and Procedure}
\label{appendix:flow_matching_training_objective_and_procedure}

The multi-modal flow matching formulation defines a training objective that jointly optimizes the continuous and discrete components of the molecular generative process. The model is parameterized by a neural network $f_{\theta}$ that maps a molecular state at time $t$ to a prediction of the final state,
\begin{equation}
\hat{G}_1(G_t, t) = f_{\theta}(G_t, t)
= \big(\hat{X}_1, \hat{A}_1, \hat{C}_1, \hat{E}_1\big).
\end{equation}
The input to the model is the current molecular state $G_t$ and time $t$, and the output consists of predictions for all modalities at the terminal time $t=1$.

\paragraph{Training pipeline.}
Each training step consists of the following procedure:

\begin{enumerate}
    \item A target molecular graph $G_1 = (X_1, A_1, C_1, E_1)$ is sampled from the data distribution.
    \item An initial state $G_0 = (X_0, A_0, C_0, E_0)$ is sampled from the prior distribution, where $X_0$ is drawn from a Gaussian distribution and all discrete variables are set to the masking state.
    \item A time variable $t \sim \mathcal{U}(0,1)$ is sampled uniformly.
    \item An intermediate state $G_t = (X_t, A_t, C_t, E_t)$ is constructed using the trajectories defined in Sections~\ref{app:continuous_flow} and~\ref{app:discrete_flow}.
    \item The model predicts the endpoint $\hat{G}_1(G_t, t)$.
\end{enumerate}

\paragraph{Continuous loss.}
For atomic coordinates, the training objective minimizes the squared error between the predicted endpoint $\hat{X}_1$ and the true coordinates $X_1$,
\begin{equation}
\mathcal{L}_{X}
=
\mathbb{E}_{t, (G_0,G_1), G_t}
\left[
w(t)\,
\| \hat{X}_1(G_t, t) - X_1 \|^2
\right],
\end{equation}
where $w(t)$ is a time-dependent weighting function.

\paragraph{Discrete loss.}
For discrete modalities, the objective is defined as a cross-entropy loss over the predicted endpoint distributions. The loss is applied only to elements that remain in the masking state at time $t$,
\begin{equation}
\mathcal{L}_{m}
=
\mathbb{E}_{t, (G_0,G_1), G_t}
\left[
- \sum_{i}
\log p_{\theta}(m_i^{(1)} \mid G_t, t)\,
\mathbf{1}(m_i^{(t)} = M)
\right],
\end{equation}
where $m \in \{A, C, E\}$ denotes a discrete modality.

\paragraph{Joint objective.}
The overall training objective is defined as a weighted sum of losses across all modalities,
\begin{equation}
\mathcal{L}
=
\lambda_X \mathcal{L}_X
+
\lambda_A \mathcal{L}_A
+
\lambda_C \mathcal{L}_C
+
\lambda_E \mathcal{L}_E,
\label{eq:joint_loss}
\end{equation}
where $\lambda_m$ are modality-specific weights that balance the contribution of each component.

\paragraph{Summary.}
The training process optimizes a single neural network to predict the final molecular state from intermediate states sampled along time-dependent trajectories. The continuous component is trained via regression toward target coordinates, while the discrete components are trained via likelihood maximization of categorical variables. This unified objective enables the model to learn a coherent generative process over molecular graphs, capturing both geometric and chemical structure within a single framework.

\subsection{Discrete CTMC Transition Derivation}
\label{appendix:Discrete_CTMC_Transition_Derivation}

This section provides additional details for the discrete CTMC transition likelihood used in RL optimization. The likelihood is defined over the same stochastic transition process used during generation, ensuring consistency between rollout and replay.

\paragraph{Factorized stochastic transition.}
For a masked token, the transition process consists of two stochastic steps. First, the process determines whether the current token should be unmasked. Second, if unmasking occurs, a target token is sampled from the predicted endpoint distribution. Let \(u_t \in \{0,1\}\) denote the unmasking decision. The resulting transition can therefore be factorized as
\begin{equation}
p(u_t,\hat z_1 \mid G_t,t)
=
p(u_t \mid G_t,t)
\,\,
p(\hat z_1 \mid u_t,G_t,t),
\label{eq:appendix_factorized_transition}
\end{equation}
where \(\hat z_1\) denotes the sampled endpoint token. Since the endpoint token only affects the transition when \(u_t=1\), the corresponding log-likelihood becomes
\begin{equation}
\log \pi_\theta
=
\log p(u_t \mid G_t,t)
+
\mathbb{I}[u_t=1]
\log p_\theta(\hat z_1 \mid G_t,t).
\label{eq:appendix_factorized_logprob}
\end{equation}

\paragraph{Non-masked transition.}
For non-masked states, the transition process only determines whether the current token should be remasked. Let \(v_t \in \{0,1\}\) denote the masking decision. The transition likelihood is therefore written as
\begin{equation}
\log \pi_\theta
=
\log p(v_t \mid G_t,t).
\label{eq:appendix_nonmasked_logprob}
\end{equation}

For discrete molecular elements indexed by $i\in\mathcal I$, the joint discrete log-likelihood is $\log\pi_\theta^{\mathrm{disc}}=\sum_{i\in\mathcal I}\log\pi_{\theta,i}$. Both masking and unmasking probabilities are clipped to $[0,1]$.

\paragraph{Replay consistency for optimization.}
During optimization, replay evaluates the likelihood under the same stochastic transition process used during rollout generation, including both masking decisions and endpoint token sampling. This ensures consistency between rollout generation and transition likelihood evaluation during RL optimization.

\subsection{Algorithm and Theoretical Analysis for Temporal Selection}
\label{appendix:Adaptive_Stochasticity_analysis}

\subsubsection{Algorithm for Temporal Selection}

Algorithm~\ref{alg:window} summarizes the temporal selection procedure.

\begin{algorithm}[htbp]
\caption{Temporal Selection}
\label{alg:window}
\begin{algorithmic}[1]

\STATE \textbf{Input:} flow matching model $\theta_{\mathrm{old}}$; integration indices $\{0,\dots,T-1\}$
\STATE \hspace{1.45cm} step size $\Delta t$; objective number $K$
\STATE \hspace{1.45cm} probe number $N_{\mathrm{probe}}$; probe stride $\mathrm{stride}$
\STATE \hspace{1.45cm} branch number $B$; probe noise $\sigma_{\mathrm{probe}}$
\STATE \textbf{Output:} temporal exploration window $\mathcal{W}$

\STATE $\mathcal{T}_{\mathrm{sub}}
\leftarrow
\{0,\mathrm{stride},2\mathrm{stride},\dots\}
\cap
\{0,\dots,T-1\}$
\hfill
\(\triangleright\) subsampled probe indices

\FORALL{$t_\star \in \mathcal{T}_{\mathrm{sub}}$}

    \STATE $\mathcal{D}_{t_\star} \leftarrow \varnothing$
    \hfill
    \(\triangleright\) store sensitivity values at \(t_\star\)

    \FOR{$i \leftarrow 1$ \textbf{to} $N_{\mathrm{probe}}$}

        \STATE $G \sim p_0$
        \hfill
        \(\triangleright\) initialize full molecular state

        \FOR{$t \leftarrow 0$ \textbf{to} $t_\star-1$}
            \STATE $G \leftarrow \mathrm{Step}_{\theta_{\mathrm{old}}}(G,t,\Delta t;\mathrm{det})$
            \hfill
            \(\triangleright\) deterministic update before branch time
        \ENDFOR

        \STATE $G_{t_\star}^{(i)} \leftarrow G$
        \hfill
        \(\triangleright\) intermediate state at probe time

        \FOR{$b \leftarrow 1$ \textbf{to} $B$}

            \STATE $G^{(i,b)} \leftarrow
            \mathrm{StochStep}_{\theta_{\mathrm{old}}}
            (G_{t_\star}^{(i)},t_\star,\Delta t;\sigma_{\mathrm{probe}})$
            \hfill
            \(\triangleright\) continuous-discrete stochastic transition

            \STATE \textcolor[HTML]{00733E}{\texttt{/* coordinates receive Gaussian noise; discrete attributes sample CTMC mask/unmask transitions */}}

            \FOR{$t \leftarrow t_\star+1$ \textbf{to} $T-1$}
                \STATE $G^{(i,b)} \leftarrow
                \mathrm{Step}_{\theta_{\mathrm{old}}}(G^{(i,b)},t,\Delta t;\mathrm{gen})$
                \hfill
                \(\triangleright\) rollout under the generation process
            \ENDFOR

            \STATE $G_1^{(i,b)} \leftarrow G^{(i,b)}$
            \hfill
            \(\triangleright\) endpoint molecule from branch \(b\)

            \FOR{$k \leftarrow 1$ \textbf{to} $K$}
                \STATE $r_{i,t_\star}^{(b,k)}
                \leftarrow
                R_k(G_1^{(i,b)})$
                \hfill
                \(\triangleright\) evaluate objective \(k\)
            \ENDFOR

        \ENDFOR

        \FOR{$k \leftarrow 1$ \textbf{to} $K$}
            \STATE $\Delta_{i,k}(t_\star)
            \leftarrow
            \max_b r_{i,t_\star}^{(b,k)}
            -
            \min_b r_{i,t_\star}^{(b,k)}$
            \hfill
            \(\triangleright\) reward sensitivity across branches

            \STATE $\mathcal{D}_{t_\star}
            \leftarrow
            \mathcal{D}_{t_\star}
            \cup
            \{(k,\Delta_{i,k}(t_\star))\}$
            \hfill
            \(\triangleright\) collect objective-wise sensitivity
        \ENDFOR

    \ENDFOR

\ENDFOR

\STATE $\mathcal{W} \leftarrow \emptyset$
\hfill
\(\triangleright\) initialize selected window

\FOR{$k \leftarrow 1$ \textbf{to} $K$}

    \FORALL{$t_\star \in \mathcal{T}_{\mathrm{sub}}$}

        \STATE $\mu_k(t_\star)
        \leftarrow
        \mathrm{mean}_i\big(\Delta_{i,k}(t_\star)\big)$
        \hfill
        \(\triangleright\) empirical sensitivity mean

        \STATE $\mathrm{SE}_k(t_\star)
        \leftarrow
        \mathrm{std}_i\big(\Delta_{i,k}(t_\star)\big) / \sqrt{N_{\mathrm{probe}}}$
        \hfill
        \(\triangleright\) standard error across probes

        \STATE $\mathrm{LCB}_k(t_\star)
        \leftarrow
        \mu_k(t_\star)
        -
        1.96\,\mathrm{SE}_k(t_\star)$
        \hfill
        \(\triangleright\) confidence-adjusted sensitivity

        \IF{$\mathrm{LCB}_k(t_\star) > 0$}
            \STATE $\mathcal{W}
            \leftarrow
            \mathcal{W}
            \cup
            \{t_\star\}$
            \hfill
            \(\triangleright\) retain reward-sensitive index
        \ENDIF

    \ENDFOR

\ENDFOR

\STATE \textbf{return} $\mathcal{W}$

\end{algorithmic}
\end{algorithm}

\subsubsection{LCB Selects Timesteps with Reliable Reward Sensitivity}

The sensitivity estimator \(\Delta_{i,k}(t)\) is non-negative but noisy because reward values depend on stochastic molecular generation dynamics. For a fixed timestep \(t\), we estimate the expected reward sensitivity for objective \(k\) using the empirical mean
\[
\mu_k(t)
=
\frac{1}{N_{\mathrm{probe}}}
\sum_{i=1}^{N_{\mathrm{probe}}}
\Delta_{i,k}(t),
\]
together with the corresponding standard error
\[
\mathrm{SE}_k(t)
=
\frac{
\mathrm{std}_i(\Delta_{i,k}(t))
}{
\sqrt{N_{\mathrm{probe}}}
}.
\]

Under mild regularity assumptions with finite variance, the sample mean satisfies asymptotic normality,
\[
\frac{
\mu_k(t)-\mathbb E[\Delta_k(t)]
}{
\mathrm{SE}_k(t)
}
\;\xrightarrow{d}\;
\mathcal N(0,1).
\]

Based on this approximation, the lower confidence bound
\[
\mathrm{LCB}_k(t)
=
\mu_k(t)-1.96\,\mathrm{SE}_k(t)
\]
provides a conservative estimate of timestep sensitivity. Timesteps with positive LCB scores are therefore more likely to correspond to reward-sensitive regions with consistently non-negligible perturbation effects.

Compared with heuristic strategies such as selecting the top percentile of timesteps, the LCB criterion provides a confidence-adjusted and distribution-aware selection rule for temporal exploration.

\subsubsection{Union Across Objectives Supports Multi-Objective Exploration}

Different objectives may exhibit distinct temporal sensitivity patterns. For objective \(k\), let \(\mathcal T_k\) denote the set of timesteps selected using the LCB criterion. The final temporal exploration window is constructed using the union
\[
\mathcal T_{\mathrm{union}}
=
\bigcup_{k=1}^{K}
\mathcal T_k .
\]

This design ensures that temporally influential regions are preserved for all objectives simultaneously. As the number of objectives increases, the union operation naturally extends the exploration window to cover all reward-sensitive regions identified across objectives, enabling stochastic exploration to adapt to multi-objective optimization settings.

\subsection{Cross-Reward Advantage Calibration}
\label{appendix:crac}
Let $r_{i,k}\in[0,1]$ be the cutoff-based reward for molecule $i$ and objective $k$, and define $v_{i,k}=1-r_{i,k}$ and $\bar v_k=N^{-1}\sum_i v_{i,k}$. For each objective, the standardized advantage and effective weight are
\begin{equation}
a_{i,k}=\operatorname{clip}\!\left(\frac{\bar v_k-v_{i,k}}{\max(\operatorname{std}_i(v_{i,k}),\epsilon_v)},-c,c\right),\qquad
\widetilde w_k=\frac{w_k\sqrt{\bar v_k+\epsilon_w}}{\sum_j w_j\sqrt{\bar v_j+\epsilon_w}}.
\end{equation}
Here, $\epsilon_v$ and $\epsilon_w$ are stability constants, and $c$ bounds the per-objective advantage. Define $z_{i,k}=\widetilde w_k a_{i,k}$, and combine positive and negative contributions as
\begin{equation}
\begin{aligned}
P_i&=\sum_k\max(z_{i,k},0), & N_i&=\sum_k\max(-z_{i,k},0),\\
\lambda_K&=\max(1,K-1), & d_i&=\tanh(P_i-\lambda_K N_i).
\end{aligned}
\end{equation}
Here $K$ is the number of objectives and $\lambda_K$ controls the weight of negative contributions. The joint-pass indicator and its batch mean are
\begin{equation}
J_i=\mathbf{1}\!\left[r_{i,k}\ge 1-\epsilon_J\ \text{for all }k\right],\qquad \bar J=\frac1N\sum_iJ_i.
\end{equation}
Here $\epsilon_J$ is the numerical tolerance for joint satisfaction. The raw and final advantages are
\begin{equation}
A_i^{\mathrm{raw}}=d_i+\beta(J_i-\bar J),\qquad
A_i^{\mathrm{CRAC}}=\frac{A_i^{\mathrm{raw}}-\frac1N\sum_j A_j^{\mathrm{raw}}}{\sqrt{1+\beta^2\bar J(1-\bar J)}}.
\end{equation}
The dense term combines objective-wise advantages and assigns a larger penalty to negative contributions. The joint-pass term prioritizes molecules satisfying all target criteria. Here $\beta$ is the joint-pass signal weight. Because $d_i\in[-1,1]$, choosing $\beta>2$ ensures that a joint-pass molecule has a higher raw advantage than a non-joint-pass molecule in the same batch. The resulting advantage is centered and scaled before the PPO update. Algorithm~\ref{alg:crac} summarizes the procedure.
\begin{algorithm}[htbp]
\caption{Cross-Reward Advantage Calibration}
\label{alg:crac}
\begin{algorithmic}[1]
\STATE \textbf{Input:} rewards $r_{i,k}$ for $i=1,\ldots,N$, $k=1,\ldots,K$; weights $w_k$; parameters $\epsilon_v,\epsilon_w,c,\epsilon_J,\beta$
\STATE \textbf{Output:} advantages $A_i^{\mathrm{CRAC}}$
\STATE $v_{i,k}\leftarrow1-r_{i,k}$; $\bar v_k\leftarrow N^{-1}\sum_i v_{i,k}$
\STATE $a_{i,k}\leftarrow\operatorname{clip}((\bar v_k-v_{i,k})/\max(\operatorname{std}_i(v_{i,k}),\epsilon_v),-c,c)$
\STATE $\widetilde w_k\leftarrow w_k\sqrt{\bar v_k+\epsilon_w}/\sum_j w_j\sqrt{\bar v_j+\epsilon_w}$; $z_{i,k}\leftarrow\widetilde w_k a_{i,k}$
\STATE $P_i\leftarrow\sum_k\max(z_{i,k},0)$; $N_i\leftarrow\sum_k\max(-z_{i,k},0)$
\STATE $d_i\leftarrow\tanh(P_i-\max(1,K-1)N_i)$
\STATE $J_i\leftarrow\mathbf1[r_{i,k}\ge1-\epsilon_J\ \text{for all }k]$; $\bar J\leftarrow N^{-1}\sum_iJ_i$
\STATE $A_i^{\mathrm{raw}}\leftarrow d_i+\beta(J_i-\bar J)$
\STATE $A_i^{\mathrm{CRAC}}\leftarrow(A_i^{\mathrm{raw}}-N^{-1}\sum_j A_j^{\mathrm{raw}})/\sqrt{1+\beta^2\bar J(1-\bar J)}$
\STATE \textbf{return} $\{A_i^{\mathrm{CRAC}}\}_{i=1}^{N}$
\end{algorithmic}
\end{algorithm}

\subsection{Overall Algorithm}

Algorithm~\ref{alg:overall} summarizes the overall algorithm.

\begin{algorithm}[htbp]
\caption{RL-Guided Continuous-Discrete Flow Matching}
\label{alg:overall}
\begin{algorithmic}[1]
\STATE \textbf{Input:} pretrained model $\theta_0$; PPO iterations $I$; window refresh frequency $f$; reuse sweeps $R$; PPO clip $\epsilon$
\STATE \textbf{Initialize:} $\theta \leftarrow \theta_0$, \ $\mathcal{W}\leftarrow\varnothing$

\FOR{$\mathrm{iter}=1$ to $I$}
    \STATE $\theta_{\mathrm{old}}\leftarrow\theta$

    \IF{$(\mathrm{iter}=1)$ or $(\mathrm{iter}\bmod f=0)$}
        \STATE $\mathcal{W}\leftarrow \textsc{TemporalSelection}(\theta_{\mathrm{old}})$ \hfill $\triangleright$ Algorithm~\ref{alg:window}
    \ENDIF

    \STATE Sample trajectories $\tau_i=\{G_t^{(i)}\}_{t=0}^{T}$ under $\theta_{\mathrm{old}}$ \hfill $\triangleright$ SDE only for $t\in\mathcal{W}$
    \STATE Evaluate molecular properties and compute reward
    \STATE Compute calibrated advantages $A_i^{\mathrm{CRAC}}$ \hfill $\triangleright$ Algorithm~\ref{alg:crac}

    \FOR{$r=1$ to $R$}
        \STATE Select update timestep set $\mathcal S$ \hfill $\triangleright$ reuse with timestep resampling
        \STATE Replay selected transitions
        \STATE Compute likelihood ratios using old and new transition likelihoods
        \STATE Update $\theta$ using the PPO objective in Eq.~\ref{eq:rl_loss}
    \ENDFOR

\ENDFOR
\end{algorithmic}
\end{algorithm}

\section{Additional Experimental Details}
\label{appendix:Additional_Experimental_Details}
\subsection{Data Sources and Descriptions}
\label{appendix:Data_Sources_and_Descriptions}
\subsubsection{Small Molecule Datasets}
\label{a ppendix:small_molecule_datasets}

We use two molecular datasets, QM9 and GEOM-Drugs, to train our flow matching models. Both datasets are widely used in 3D de novo molecular design and are commonly treated as standard benchmarks in this area. Table~\ref{tab:datasets} summarizes the data sources along with brief descriptions of each dataset.

\begin{table}[htbp]
\centering
\caption{Summary of QM9 and GEOM-Drugs datasets}
\label{tab:datasets}
\small
\begin{tabularx}{\textwidth}{>{\raggedright\arraybackslash}p{0.22\textwidth}
                            >{\raggedright\arraybackslash}X
                            >{\raggedright\arraybackslash}X}
\toprule
\textbf{Items} & \textbf{QM9} & \textbf{GEOM-Drugs} \\
\midrule

Description 
& 3D small organic molecules dataset from the GDB-17 chemical space 
& Large-scale 3D dataset of drug-like molecules \\

Number of molecules 
& $\sim$134K 
& $\sim$304K \\

Properties 
& Energies, dipole moments, polarizabilities, etc. 
& N/A \\

Unique atom types 
& C, N, O, F, H 
& C, H, N, O, F, P, S, Cl, Br, I \\

Atom valency / states 
& Simple, fixed valency patterns 
& Diverse valency and charge states across atom types \\

Molecular size 
& Small molecules ($\leq$ 9 heavy atoms) 
& Larger, more complex drug-like molecules \\

Data complexity 
& Relatively simple and clean 
& More complex and diverse \\

Sources 
& \url{http://deepchem.io.s3-website-us-west-1.amazonaws.com/datasets/gdb9.tar.gz} 
& \url{https://bits.csb.pitt.edu/files/geom_raw/} \\

\bottomrule
\end{tabularx}
\end{table}

\subsubsection{Target Proteins}
\label{a ppendix:target_proteins}

In this study, we select two representative target proteins, EGFR and Pin1, and aim to design small-molecule inhibitors with high binding affinity toward these targets. These two proteins are chosen because they exhibit substantial differences in biological function, binding characteristics, and druggability, allowing us to evaluate the performance of our method under different binding mechanisms and levels of design difficulty.

As summarized in Table~\ref{tab:proteins}, both EGFR and Pin1 are important targets in oncology, particularly in lung and breast cancers. The ATP-binding pocket of EGFR is relatively well-defined and has been widely explored in prior studies, with some inhibitors reported. EGFR is generally considered a relatively tractable target for drug design. In contrast, Pin1 presents a significantly higher level of difficulty. Pin1 contains a shallow and highly dynamic catalytic pocket, where ligand binding often depends on subtle conformational changes and stereochemical matching rather than strong and well-defined local interactions. This makes the design of high-affinity ligands for Pin1 more challenging.

\begin{table}[htbp]
\centering
\caption{Comparison of EGFR and Pin1}
\label{tab:proteins}
\small
\begin{tabularx}{\textwidth}{>{\raggedright\arraybackslash}p{0.22\textwidth}
                            >{\raggedright\arraybackslash}X
                            >{\raggedright\arraybackslash}X}
\toprule
\textbf{Items} & \textbf{EGFR (PDB ID: 3W2S)} & \textbf{Pin1 (PDB ID: 3IKD)} \\
\midrule
Protein type 
& Tyrosine kinase 
& Peptidyl-prolyl isomerase \\

Structural domain 
& Kinase domain 
& Catalytic domain \\

Pocket exposure 
& More enclosed 
& More solvent-exposed \\

Resolution 
& 1.90 \AA 
& 2.0 \AA \\

Binding pocket 
& Deep, well-defined ATP-binding pocket 
& Shallow, flexible catalytic pocket \\

Related diseases 
& Non-Small Cell Lung Cancer (NSCLC), glioblastoma 
& Triple-Negative Breast Cancer (TNBC), hepatocellular carcinoma \\

Druggability 
& Relatively easier 
& More challenging \\

Sources 
& \url{https://www.rcsb.org/structure/3W2S}
& \url{https://www.rcsb.org/structure/3IKD} \\
\bottomrule
\end{tabularx}
\end{table}

For structural selection, we use 3W2S and 3IKD as the crystal structures of EGFR and Pin1, respectively. The 3W2S structure corresponds to the kinase domain of EGFR, where the ATP-binding pocket is well-defined and relatively deep, facilitating ligand entry and stable interactions. In contrast, 3IKD corresponds to the catalytic domain of Pin1, where the binding site is more shallow and solvent-exposed, with more limited interaction patterns, making stable binding more difficult. Both structures are high-resolution protein–ligand co-crystal structures resolved by X-ray crystallography and are widely used in molecular discovery tasks. Their high resolution supports stable docking with Vina, and the presence of co-crystallized ligands clearly defines the binding pocket, providing a reference for constructing the docking box.

\subsection{Data Preprocessing and Splitting}
\label{appendix:Data_Preprocessing_and_Splitting}
\subsubsection{Molecular Data}
\textbf{Processing.} QM9 and GEOM-Drugs are processed using a consistent pipeline to obtain valid 3D molecular structures for pretraining the flow matching models. For each molecule, atomic coordinates, atom types, formal charges, and bond information are extracted from the raw data. Molecules that cannot be successfully parsed or processed are removed, including cases with missing or invalid 3D coordinates, chemically invalid structures, or errors encountered during data cleaning. As a result, only chemically valid and well-defined molecular structures are retained.

For the GEOM-Drugs dataset, each molecule is associated with multiple 3D conformations corresponding to different low-energy geometries of the same molecular graph. These conformations are treated as independent samples, as they capture structural variability while preserving the underlying chemical identity. When multiple conformations are available, a subset may be selected per molecule to control the overall dataset size. This formulation allows the model to learn distributions over 3D molecular geometries rather than a single fixed structure per molecule. For the QM9 dataset, molecules are directly loaded from the original SDF files with explicit hydrogens retained. Uncharacterized molecules provided in the official QM9 filtering list are removed prior to processing.

\begin{figure}[htbp]
\centering
\includegraphics[width=0.99\linewidth]{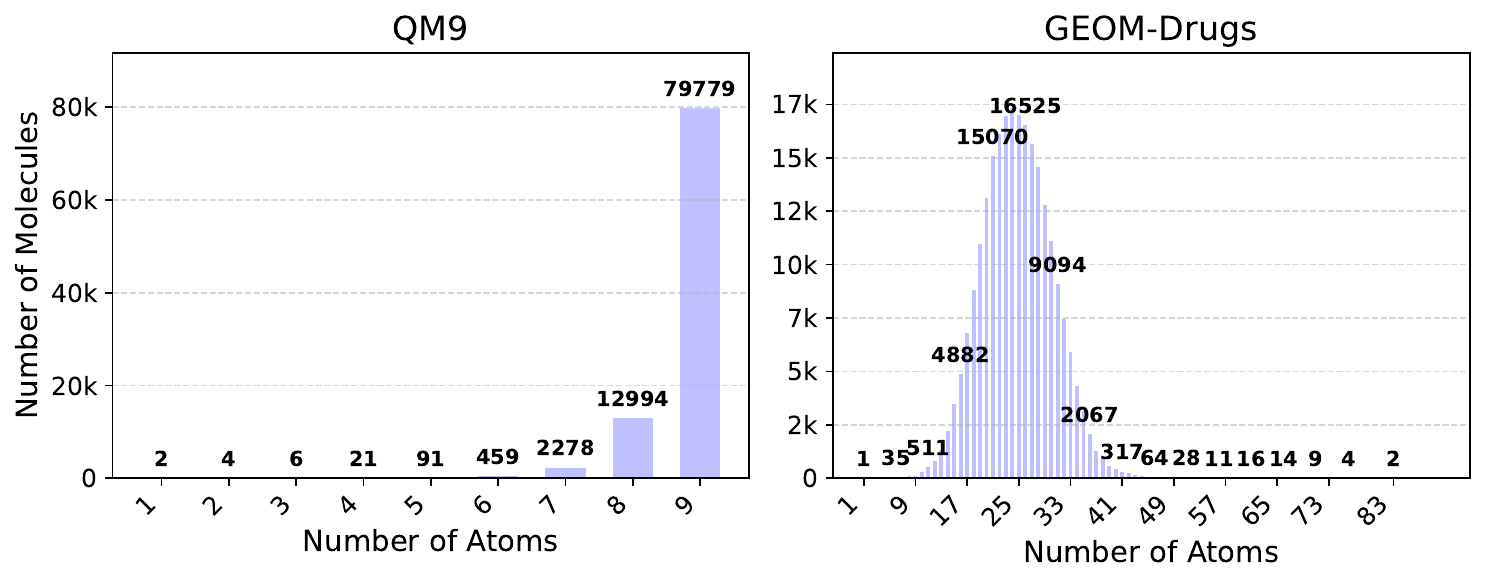}
\caption{Distribution of molecules by number of heavy atoms (hydrogens removed)}
\label{fig:atom_dist}
\end{figure}

\textbf{Splitting.} The datasets are randomly shuffled at the molecule level and divided into training (80\%), validation (10\%), and test (10\%) sets. For GEOM-Drugs, all conformations belonging to the same molecule are assigned to the same split to prevent information leakage across splits. The distribution of molecular sizes in training data is reported in Fig.~\ref{fig:atom_dist}.

\subsubsection{Target Protein Data}
After obtaining the protein–ligand co-crystal structures, the receptor structures are prepared following standard preprocessing procedures. The target protein chain (e.g., chain A) is retained, while crystallographic water molecules, ions, buffer components, and other non-relevant small molecules are removed. The co-crystallized ligand is extracted for subsequent redocking validation. The protein structure is then inspected for structural consistency, including missing residues or abnormal atoms, and hydrogen atoms are added, with particular attention to polar hydrogens. Partial charges are assigned using the Gasteiger scheme, and the processed receptor is converted into the PDBQT format required for docking.

\subsection{Docking Protocol}
\label{appendix:Docking_Protocol}
We use QuickVina2-GPU-2.1 as the molecular docking engine. After identifying the ligand from the protein structure, the binding region required by Vina is defined automatically based on its position, following Alg.~\ref{alg:binding_box}. The docking box center is set to the center of the ligand bounding box, computed from the minimum and maximum ligand atomic coordinates along each spatial dimension. The box size is determined from the ligand spatial extent with additional padding to ensure sufficient coverage of the binding pocket while avoiding an excessively large search region.

\begin{algorithm}[t]
\centering
\caption{Ligand-Guided Docking Box Identification}
\label{alg:binding_box}
\begin{algorithmic}[1]
\REQUIRE Protein structure file $F_{\mathrm{pdb}}$, chain identifier $c$, optional ligand residue index $r_{\mathrm{lig}}$, padding $p$, box size constraints $\mathbf{s}_{\min}$ and $\mathbf{s}_{\max}$.
\ENSURE Docking box center $\mathbf{c}_{\mathrm{box}} \in \mathbb{R}^{3}$, box size $\mathbf{s}_{\mathrm{box}} \in \mathbb{R}^{3}$.

\STATE \textcolor{mygreen}{/* Step 1: Protein Chain Selection */}
\STATE Load structure $U \leftarrow F_{\mathrm{pdb}}$
\STATE $P \leftarrow \mathrm{SelectProtein}(U)$ \hfill $\triangleright$ extract all protein atoms from the structure
\IF{$c$ is provided}
    \STATE $P_c \leftarrow \mathrm{SelectChain}(P,c)$ \hfill $\triangleright$ restrict the protein to the specified chain
    \IF{$P_c \neq \emptyset$}
        \STATE $P \leftarrow P_c$
    \ENDIF
\ENDIF

\STATE \textcolor{mygreen}{/* Step 2: Candidate Ligand Detection */}
\STATE $\mathcal{C} \leftarrow \emptyset$ \hfill $\triangleright$ initialize candidate ligand set
\FOR{each residue $R$ in $U$}
    \IF{$R$ is a standard residue or solvent/ion}
        \STATE \textbf{continue} \hfill $\triangleright$ remove protein residues and non-ligand molecules
    \ENDIF
    \IF{$|R| < 5$}
        \STATE \textbf{continue} \hfill $\triangleright$ discard small fragments or isolated hetero atoms
    \ENDIF
    \STATE $d_R \leftarrow \min_{\mathbf{x}\in R,\,\mathbf{y}\in P}\|\mathbf{x}-\mathbf{y}\|_2$ \hfill $\triangleright$ minimum atom-wise ligand-protein distance
    \IF{$d_R \leq 6.0$}
        \STATE $\mathcal{C} \leftarrow \mathcal{C} \cup \{(R,d_R)\}$ \hfill $\triangleright$ keep candidates close to the binding region
    \ENDIF
\ENDFOR
\IF{$\mathcal{C}=\emptyset$}
    \STATE Raise error: no ligand found
\ENDIF

\STATE \textcolor{mygreen}{/* Step 3: Reference Ligand Selection */}
\IF{$r_{\mathrm{lig}}$ is provided}
    \STATE $L \leftarrow \{R \mid (R,d_R)\in\mathcal{C},\ \mathrm{resid}(R)=r_{\mathrm{lig}}\}$ \hfill $\triangleright$ explicitly select the specified ligand
    \IF{$L=\emptyset$}
        \STATE Raise error: ligand residue index not found
    \ENDIF
\ELSE
    \STATE $L \leftarrow \arg\min_{(R,d_R)\in\mathcal{C}} d_R$ \hfill $\triangleright$ automatically choose the closest ligand
\ENDIF

\STATE \textcolor{mygreen}{/* Step 4: Docking Box Construction */}
\STATE $\mathbf{X}_L \leftarrow \mathrm{Coordinates}(L)$ \hfill $\triangleright$ extract 3D coordinates of ligand atoms
\STATE $\mathbf{x}_{\min} \leftarrow \min(\mathbf{X}_L)$ \hfill $\triangleright$ coordinate-wise lower corner of ligand bounding box
\STATE $\mathbf{x}_{\max} \leftarrow \max(\mathbf{X}_L)$ \hfill $\triangleright$ coordinate-wise upper corner of ligand bounding box
\STATE $\mathbf{c}_{\mathrm{box}} \leftarrow (\mathbf{x}_{\min}+\mathbf{x}_{\max})/2$ \hfill $\triangleright$ center of the ligand bounding box
\STATE $\mathbf{s}_{\mathrm{box}} \leftarrow (\mathbf{x}_{\max}-\mathbf{x}_{\min}) + 2p$ \hfill $\triangleright$ add symmetric padding along each dimension
\STATE $\mathbf{s}_{\mathrm{box}} \leftarrow \max(\mathbf{s}_{\mathrm{box}},\mathbf{s}_{\min})$ \hfill $\triangleright$ enforce minimum search-space size
\STATE $\mathbf{s}_{\mathrm{box}} \leftarrow \min(\mathbf{s}_{\mathrm{box}},\mathbf{s}_{\max})$ \hfill $\triangleright$ limit excessive docking-space volume

\STATE \RETURN $\mathbf{c}_{\mathrm{box}}, \mathbf{s}_{\mathrm{box}}$
\end{algorithmic}
\end{algorithm}

\begin{figure}[htbp]
\centering
\includegraphics[width=0.99\linewidth]{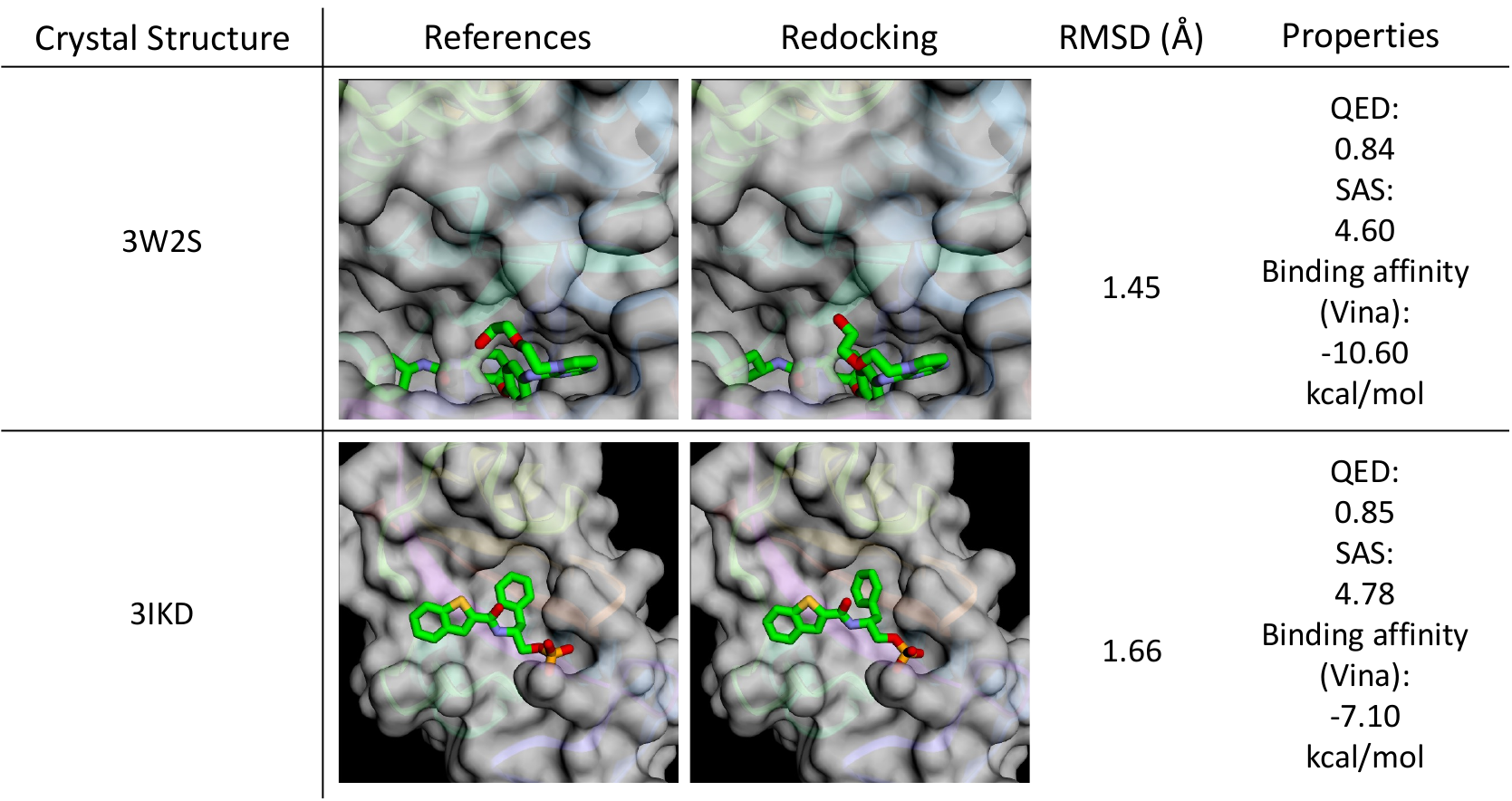}
\caption{Redocking Validation on EGFR (PDB ID: 3W2S) and Pin1 (PDB ID: 3IKD)}
\label{fig:redocking}
\end{figure}

With the binding box and center determined, redocking is performed to validate the docking setup. Using search depths of 8 for 3W2S and 28 for 3IKD, the resulting docked poses are consistent with the crystallographic binding mode, as illustrated in Fig.~\ref{fig:redocking}. To further quantify this agreement, the Root-Mean-Square Deviation (RMSD) of atomic positions is computed between the docked pose and the reference ligand. RMSD is a standard metric for evaluating docking accuracy, where values below 2.0 Å are generally considered to indicate successful recovery of the binding pose. The obtained RMSD values (3W2S: 1.45 Å, 3IKD: 1.66 Å) fall within the acceptable range, confirming that the docking protocol reliably reproduces experimentally observed binding configurations. It is worth noting that the search depth directly affects computational cost, with larger values generally improving docking accuracy at the expense of increased runtime. While higher search depths can lead to more accurate pose estimation, they are less suitable for large-scale or repeated docking tasks. Therefore, the selected values provide a practical trade-off between accuracy and efficiency. In addition, the physicochemical properties of the reference ligands are evaluated, including QED, SAS, and binding affinity estimated by Vina. These values are reported as references.

\subsection{Ranges, Normalization, and Scaling}
\label{appendix:Ranges_Normalization_and_Scaling}
In our experiments, the considered molecular properties differ in both scale and direction (i.e., some properties are maximized while others are minimized). To achieve consistent comparison and avoid numerical discrepancies, thereby enabling stable optimization, all properties are rescaled to the range (0, 1) and unified to the same direction. Tables~\ref{tab:geom_property_normalization} and~\ref{tab:qm9_property_normalization} summarize the rescaling strategies, value ranges, and directionality of the properties used in the GEOM-Drugs and QM9 datasets, respectively.

\begin{table}[htbp]
\centering
\caption{Drug property normalization strategies.}
\label{tab:geom_property_normalization}
\small
\begin{tabularx}{\textwidth}{lcc>{\raggedright\arraybackslash}Xcc}
\toprule
\textbf{Property} 
& \textbf{Range} 
& \textbf{Direction} 
& \textbf{Normalization} 
& \textbf{Adjusted Range} 
& \textbf{Adjusted Direction} \\
\midrule

QED 
& $(0,1)$ 
& $\uparrow$ 
& N/A 
& N/A 
& N/A \\

SAS 
& $(1,10)$ 
& $\downarrow$ 
& $\hat{s}_{\mathrm{sas}} = 1 - 
\frac{\mathrm{clip}(s_{\mathrm{sas}} - s_{\mathrm{sas}}^{\min},\, 0,\, s_{\mathrm{sas}}^{\max} - s_{\mathrm{sas}}^{\min})}
{s_{\mathrm{sas}}^{\max} - s_{\mathrm{sas}}^{\min}}$
& $(0,1)$ 
& $\uparrow$ \\

Affinity 
& $(-12,0)$ 
& $\downarrow$ 
& $\hat{s}_{\mathrm{aff}} = 1 - 
\frac{\mathrm{clip}(s_{\mathrm{aff}} - s_{\mathrm{aff}}^{\min},\, 0,\, s_{\mathrm{aff}}^{\max} - s_{\mathrm{aff}}^{\min})}
{s_{\mathrm{aff}}^{\max} - s_{\mathrm{aff}}^{\min}}$
& $(0,1)$ 
& $\uparrow$ \\

\bottomrule
\end{tabularx}

\raggedright
\vspace{0.2em}
\footnotesize{Note: 
$\uparrow$ indicates that larger values are preferred, while $\downarrow$ indicates that smaller values are preferred.
Here, $s$ denotes the original property value, $\hat{s}$ denotes the normalized property value,
and $s^{\min}$ and $s^{\max}$ denote the lower and upper bounds of the corresponding property, respectively.
}
\end{table}
\begin{table}[htbp]
\centering
\caption{QM9 property normalization strategies.}
\label{tab:qm9_property_normalization}
\small
\begin{tabularx}{\textwidth}{lcc>{\raggedright\arraybackslash}Xcc}
\toprule
\textbf{Property (Unit)} 
& \textbf{Range} 
& \textbf{Direction} 
& \textbf{Normalization} 
& \textbf{Adjusted Range} 
& \textbf{Adjusted Direction} \\
\midrule

HOMO (eV) 
& $\sim[-10, -4]$ 
& Window 
& $\hat{s}_{\mathrm{HOMO}} = r(s_{\mathrm{HOMO}})$ 
& $(0,1)$ 
& $\uparrow$ \\

LUMO (eV) 
& $\sim[-4, 2]$ 
& Window 
& $\hat{s}_{\mathrm{LUMO}} = r(s_{\mathrm{LUMO}})$ 
& $(0,1)$ 
& $\uparrow$ \\

gap (eV) 
& $\sim[1, 12]$ 
& Window 
& $\hat{s}_{\mathrm{gap}} = r(s_{\mathrm{gap}})$ 
& $(0,1)$ 
& $\uparrow$ \\

$\mu$ (Debye) 
& $\sim[0, 8]$ 
& Window 
& $\hat{s}_{\mu} = r(s_{\mu})$ 
& $(0,1)$ 
& $\uparrow$ \\

$\alpha$ ($a_0^3$) 
& $\sim[10, 200]$ 
& $\uparrow$ 
& $\hat{s}_{\alpha} = \mathrm{clip}\!\left( \frac{s_{\alpha} - s_{\alpha}^{r}}{s_{\alpha}^{s} - s_{\alpha}^{r}}, 0, 1 \right)$ 
& $(0,1)$ 
& $\uparrow$ \\

\bottomrule
\end{tabularx}

\vspace{0.5em}
\raggedright
\footnotesize{Note: $\uparrow$ indicates that larger normalized values are preferred, and ``Window'' indicates that values within a predefined desired interval are preferred. For $\alpha$, the optimization reward increases with polarizability, while the success criterion evaluates whether $\alpha$ falls within the target interval $[60, 120]$. Here, $s$ denotes the original property value and $\hat{s}$ the normalized value. For window-based properties, $r(\cdot)$ maps values inside the desired window to 1 and gradually decreases the score outside the window according to the empirical range. Specifically, $\hat{s} = 1$ if $l_d \le s \le u_d$, 
$\hat{s} = \max\!\left(0, 1 - \frac{l_d - s}{l_d - l_e}\right)$ if $s < l_d$, and $\hat{s} = \max\!\left(0, 1 - \frac{s - u_d}{u_e - u_d}\right)$ if $s > u_d$, 
where $[l_d, u_d]$ denotes the desired window and $[l_e, u_e]$ denotes the empirical range. The empirical ranges are estimated from the QM9 dataset; due to model approximation and prediction uncertainty, predicted values may occasionally fall outside these ranges.
}
\end{table}

\subsection{Baseline Implementation and Analysis}
Baselines are implemented according to their respective formulations. We use the same evaluation criteria across methods and align experimental settings where applicable. Details of the sampling-based baselines are provided below.
\label{appendix:Baseline_Implementation_and_Analysis}
\subsubsection{Best-of-N}

\textbf{Design and rationales.} Best-of-N is a sampling-based strategy. For each run, we independently sample N sets of molecules from the same model under identical settings. The model parameters are not updated, and no additional training is introduced. The final molecular set is then constructed by comparing the score of each molecule. The score is defined based on its distance to the assigned property cutoffs. Specifically, if a molecule satisfies the cutoff (strict) for a given property, it receives the full score for that property. Otherwise, its score is determined by how far it is from the cutoff, with smaller gaps leading to higher scores. All properties are normalized before computing the final combined score. The final score is calculated by summing the scores across all properties, resulting in a continuous value. This score reflects how well a molecule simultaneously satisfies the desired cutoffs.

The key idea behind this method is that it improves performance by increasing the chance of observing a high-quality sample from the same underlying model distribution. Since molecular generation is stochastic, a single sampling run may miss molecules that satisfy multiple desired property constraints. By repeating the sampling process multiple times, Best-of-N increases the chance of observing better molecules from the same model distribution. Therefore, its improvement mainly comes from sampling-based selection rather than from changing the generative model itself. In other words, it trades additional computation cost for a higher chance of generating better molecules.

\textbf{Analysis of the Choice of N.} The computational cost mainly depends on N, which denotes the number of independent sampling attempts for the same generation task. Empirically, increasing N expands the candidate space and improves the probability of obtaining high-quality molecules, which typically leads to better performance. This trend is also observed in our experiments, where the success rate consistently improves as N increases.

However, this improvement is neither linear nor sustained exponentially, and instead exhibits clear diminishing returns. As N grows, newly generated molecules may tend to be increasingly similar to existing ones in distribution, making it less likely to discover substantially better molecules. As a result, the incremental performance gain gradually narrows. In contrast, the computational cost increases linearly with N. In our setting, each additional sampling round requires a full evaluation of the generated molecules, including Vina docking, which is computationally expensive, especially when invoked frequently at scale. Therefore, as N increases, the overall cost from both generation and evaluation rises significantly.

As a result, N cannot be increased indefinitely in practice and must be chosen to balance performance gains and computational cost. Based on our results, the quality of the selected molecules improves steadily as N increases, but at larger values (e.g., N=12), the computational cost becomes substantial, requiring approximately two days of runtime. This indicates that the practical benefit of increasing N diminishes beyond a certain point. In comparison, a moderate value such as N=3 provides a more favorable trade-off between performance and efficiency, making it a more practical choice in real-world applications.

\subsubsection{SMC}
\textbf{Design and rationales.} SMC is an inference-time trajectory-based method and does not require updating model weights. SMC maintains a population of particles and propagates them through the same generative process defined by the pretrained model. Each particle corresponds to one partially generated molecular trajectory. During sampling, the particles are periodically evaluated using a scoring function similar to that used in Best-of-N.

At each resampling step, particles are resampled according to their weights and then continue the generation process from the selected intermediate states. In this way, SMC gradually shifts the particle population toward trajectories that are more likely to produce molecules satisfying the desired property requirements, while still following the original model dynamics.

The key idea behind this method is that it guides the sampling process toward better regions of the model distribution by repeatedly favoring more promising intermediate states. Since molecular generation is stochastic, sampling only once may lead many trajectories to suboptimal molecules. SMC increases the chance of obtaining high-quality molecules by allocating more sampling effort to trajectories that already show better partial structures. This allows the method to better exploit the model without modifying its parameters. By adjusting the strength of this guidance, one can control how strongly the particle selection is biased toward higher-reward candidates.

\textbf{Analysis of the choice of $\beta$.} The parameter $\beta$ controls the strength of the weighting during the resampling process. A smaller $\beta$ leads to a flatter weight distribution, where particles are sampled more uniformly, resulting in higher diversity but weaker guidance toward high-quality molecules. In contrast, a larger $\beta$ increases the emphasis on high-reward particles, making the selection more aggressive, but may cause the process to concentrate on a narrower set of trajectories, reducing diversity and leading to more repeated molecules.

This trend is clearly reflected in the presented results. As $\beta$ increases from 0.5 to 2, the success rate improves significantly, especially under the strict criterion. This indicates that stronger weighting helps concentrate the particle population on trajectories that are more likely to satisfy the desired property cutoffs. However, this comes at the cost of reduced diversity, as shown by the drop in uniqueness. This is expected, since more aggressive resampling tends to repeatedly select a smaller set of high-reward candidates. In practice, choosing a moderate value such as $\beta = 1$ provides a more balanced trade-off between maintaining diversity and improving success rate.

\subsection{Hyperparameters for Flow Matching Model Training}
\label{appendix:Hyperparameters_for_Flow_Matching_Model_Training}

Flow matching models are trained separately on the GEOM-Drugs and QM9 datasets. With reference to the original FlowMol3 implementation, the final configurations used for training are summarized in Table~\ref{tab:flow_matching_hyperparameters}.

\begin{table}[htbp]
\centering
\caption{Hyperparameters used for training on flow matching models.}
\label{tab:flow_matching_hyperparameters}
\begin{tabular}{lcc}
\toprule
\textbf{Parameter} & \textbf{GEOM-Drugs} & \textbf{QM9} \\
\midrule
Batch size & 12 & 128 \\
Num workers & 6 & 8 \\
Max epochs & 20 & 20 \\
Learning rate & $1\times10^{-4}$ & $2.5\times10^{-4}$ \\
Weight decay & $1\times10^{-12}$ & $1\times10^{-12}$ \\
Gradient accumulation & 1 & 1 \\
Max edges & 50{,}000 & 200{,}000 \\
\midrule
Hidden scalar dim & 256 & 256 \\
Edge feature dim & 128 & 128 \\
Vector channels & 32 & 16 \\
\# Molecule updates & 6 & 8 \\
Message GVP layers & 3 & 3 \\
Update GVP layers & 3 & 3 \\
RBF max distance & 10 & 14 \\
RBF dim & 32 & 32 \\
Time embedding dim & 64 & 64 \\
\midrule
Distortion prob & 0.2 & 0.2 \\
Fake atom prob & 0.3 & 0.3 \\
Self-conditioning & True & True \\
\bottomrule
\end{tabular}
\end{table}

\subsection{RL-guided Flow Matching Model Training and Hyperparameters}
\label{appendix:RL_Training_and_Hyperparameters}
In this work, RL-guided optimization of flow matching models involves two main types of tasks. For the flow matching model pretrained on GEOM-Drugs, the optimization targets are drug-relevant molecular properties, including QED, SAS, and binding affinity. For the flow matching model pretrained on QM9, the optimization targets are molecular physical properties, including HOMO, LUMO, gap, $\mu$, and $\alpha$.

For both types of tasks, we conduct an extensive exploration of key hyperparameters to identify suitable configurations for RL-guided fine-tuning. The explored hyperparameter ranges are summarized in Table~\ref{tab:rl_hyperparameter_search}. Based on this exploration, the final selected configurations for RL-guided flow matching models pretrained on each dataset are reported in Table~\ref{tab:rl_selected_hyperparameters}.

\begin{table}[htbp]
\centering
\caption{RL hyperparameter search space.}
\label{tab:rl_hyperparameter_search}
\begin{tabular}{lc}
\toprule
\textbf{Hyperparameter} & \textbf{Search Range} \\
\midrule
Batch size 
& 4, 8, 16, 32, 64, 128, 256 \\

SDE noise scale ($\sigma$) 
& 0.01, 0.03, 0.05, 0.08, 0.1, 0.2, 0.5, 1 \\

Learning rate 
& 2e-4, 1e-4, 2e-5, 5e-5, 1e-5, 5e-6, 1e-6 \\

Clipping parameter 
& 3e-3, 1e-3, 1e-4, 2e-4, 3e-4, 4e-4, 5e-4, 7e-4 \\

Gradient clipping norm 
& 0.1, 0.3, 0.5, 0.8, 1.0, 2.0, 5.0 \\

Update iterations 
& 1, 2, 3, 4, 5, 10 \\
\bottomrule
\end{tabular}
\end{table}

\begin{table}[htbp]
\centering
\caption{Final RL hyperparameter configurations.}
\label{tab:rl_selected_hyperparameters}
\begin{tabular}{lcc}
\toprule
\textbf{Hyperparameter} & \textbf{GEOM-Drugs} & \textbf{QM9} \\
\midrule
Batch size 
& 64 & 128 \\

SDE noise scale ($\sigma$) 
& 0.05 & 0.05 \\

Learning rate 
& 1e-5 & 2e-5 \\

Clipping parameter 
& 5e-4 & 3e-4 \\

Gradient clipping norm 
& 0.5 & 0.5 \\

Update iterations 
& 3 & 3 \\
\bottomrule
\end{tabular}
\end{table}

\subsection{Performance Metrics}
\label{appendix:Performance_Metrics}
\subsubsection{Standard Metrics}
The performance of the flow matching model is primarily evaluated based on the quality of the generated molecules. We adopt a set of widely used metrics in the field to assess different aspects of molecular quality, including validity, uniqueness, novelty, the proportion of molecules that are simultaneously valid, unique, and novel (VUN), atom stability, molecular stability, and success rate.

\paragraph{Validity.}
Validity measures the proportion of generated molecules that can be successfully parsed into chemically valid molecular structures. In our setting, each generated molecule is represented as a molecular graph with associated 3D coordinates, where atom types and bond structures are directly predicted by the model. The resulting graph is then converted into an RDKit molecule object for validation and standardization, including sanitization and canonical SMILES generation. A molecule is considered valid if this process completes without errors. Formally, validity is defined as
\begin{equation}
\mathrm{Validity}
=
\frac{
\left|\left\{ m \in \mathcal{M}_{\mathrm{gen}} \mid m \text{ is valid} \right\}\right|
}{
\left|\mathcal{M}_{\mathrm{gen}}\right|
}
\times 100\% ,
\end{equation}
where $\mathcal{M}_{\mathrm{gen}}$ denotes the set of all generated molecules. The value of validity ranges from $0\%$ to $100\%$, where a higher value indicates that a larger fraction of generated samples can be interpreted as chemically valid molecules. A high validity score reflects the model's ability to generate chemically meaningful structures, while a low validity score suggests failures in molecular reconstruction, invalid valence patterns, or unrealistic local geometries.

\paragraph{Uniqueness.}
Uniqueness measures the diversity of the generated valid molecules. After invalid molecules are removed, duplicate molecules are identified based on their canonical SMILES strings. The uniqueness score is then computed as the proportion of distinct molecules among all valid molecules:
\begin{equation}
\mathrm{Uniqueness}
=
\frac{
\left|\mathrm{Unique}\left(\mathcal{M}_{\mathrm{valid}}\right)\right|
}{
\left|\mathcal{M}_{\mathrm{valid}}\right|
}
\times 100\% ,
\end{equation}
where $\mathcal{M}_{\mathrm{valid}}$ denotes the set of valid generated molecules, and $\mathrm{Unique}(\cdot)$ removes duplicated molecules according to canonical SMILES representations. The value ranges from $0\%$ to $100\%$. A higher uniqueness score indicates fewer exact duplicates among valid generated molecules. In contrast, a low uniqueness score may indicate limited exploration of chemical space or mode collapse.

\paragraph{Novelty.}
Novelty quantifies the proportion of unique valid generated molecules that do not appear in the training set. This metric evaluates whether the model can generate new molecular structures rather than simply reproducing molecules observed during training. The comparison is performed using canonical SMILES strings to ensure a consistent molecular representation. Formally, novelty is defined as
\begin{equation}
\mathrm{Novelty}
=
\frac{
\left|\left\{ m \in \mathrm{Unique}\left(\mathcal{M}_{\mathrm{valid}}\right)
\mid
m \notin \mathcal{M}_{\mathrm{train}}
\right\}\right|
}{
\left|\mathrm{Unique}\left(\mathcal{M}_{\mathrm{valid}}\right)\right|
}
\times 100\% ,
\end{equation}
where $\mathcal{M}_{\mathrm{train}}$ denotes the set of molecules in the training data. The novelty score ranges from $0\%$ to $100\%$. A high novelty score indicates that generated molecules are absent from the training set under the canonical SMILES representation. A low novelty score suggests that the model may overfit to the training data or memorize known molecules.

\paragraph{VUN.}
VUN measures the proportion of generated molecules that are simultaneously valid, unique, and novel. Compared with validity, uniqueness, and novelty considered separately, VUN provides a more integrated measure of useful de novo generation quality. A molecule contributes to VUN only if it can be chemically reconstructed, is not a duplicate of another generated molecule, and does not appear in the training set. Formally, VUN is defined as
\begin{equation}
\mathrm{VUN}
=
\frac{
\left|
\left\{
m \in \mathrm{Unique}\left(\mathcal{M}_{\mathrm{valid}}\right)
\mid
m \notin \mathcal{M}_{\mathrm{train}}
\right\}
\right|
}{
\left|\mathcal{M}_{\mathrm{gen}}\right|
}
\times 100\% .
\end{equation}
The value of VUN ranges from $0\%$ to $100\%$. A high VUN score means that a large fraction of all generated outputs are meaningful de novo candidates that are chemically valid, structurally non-redundant, and absent from the training set. A low VUN score indicates that many generated samples fail due to invalidity, duplication, or overlap with the training data.

\paragraph{Atom Stability.}
Atom stability evaluates whether generated molecules satisfy local chemical constraints at the atom level. After molecular reconstruction and bond inference, each atom is checked to determine whether its inferred valence is chemically allowed. This metric reflects whether the model places atoms in chemically reasonable local bonding environments. Formally, atom stability is defined as
\begin{equation}
\mathrm{Atom\ Stability}
=
\frac{
\sum_{m \in \mathcal{M}_{\mathrm{valid}}}
\sum_{a \in \mathcal{A}(m)}
\mathbf{1}\left[a \text{ obeys valency}\right]
}{
\sum_{m \in \mathcal{M}_{\mathrm{valid}}}
\left|\mathcal{A}(m)\right|
}
\times 100\% ,
\end{equation}
where $\mathcal{A}(m)$ denotes the set of atoms in molecule $m$, and $\mathbf{1}[\cdot]$ is an indicator function. The value ranges from $0\%$ to $100\%$. A high atom stability score indicates that most generated atoms satisfy standard valence constraints, suggesting that the model respects fundamental atom-level chemical rules. A low atom stability score suggests frequent local bonding errors, such as over-bonded or under-bonded atoms.

\paragraph{Molecular Stability.}
Molecular stability measures whether a generated molecule forms a chemically coherent structure as a whole. While atom stability focuses on local atom-level valency, molecular stability evaluates whether the entire molecule is structurally plausible. A molecule is considered stable if all atoms satisfy the required valency constraints and the reconstructed molecule does not contain severe structural inconsistencies such as invalid bonding patterns or disconnected fragments. Formally, molecular stability is defined as
\begin{equation}
\mathrm{Molecular\ Stability}
=
\frac{
\left|
\left\{
m \in \mathcal{M}_{\mathrm{valid}}
\mid
m \text{ is chemically stable}
\right\}
\right|
}{
\left|\mathcal{M}_{\mathrm{valid}}\right|
}
\times 100\% .
\end{equation}
The molecular stability score ranges from $0\%$ to $100\%$. A high value indicates that most valid generated molecules form chemically coherent molecular graphs, whereas a low value suggests that many molecules contain structural discontinuities, invalid bonding configurations, or other global chemical inconsistencies.

\paragraph{Success Rate.}
In addition to general generation quality, we evaluate whether the generated molecules satisfy desired drug-like property constraints. The success rate measures the proportion of generated molecules that simultaneously meet all predefined property cutoffs. In the GEOM-Drugs experiments, we consider three key properties: QED, SAS, and binding affinity measured by Vina. QED is maximized because larger values indicate more drug-like molecules. SAS is minimized because smaller values indicate easier synthesis. Before computing the synthetic accessibility score, explicit hydrogen atoms are removed from each molecule. Binding affinity is also minimized because more negative values indicate stronger predicted binding affinity.

Given a set of property constraints $\mathcal{C}$, the success rate is defined as
\begin{equation}
\mathrm{Success\ Rate}
=
\frac{
\left|
\left\{
m \in \mathcal{M}_{\mathrm{gen}}
\mid
m \text{ satisfies all constraints in } \mathcal{C}
\right\}
\right|
}{
\left|\mathcal{M}_{\mathrm{gen}}\right|
}
\times 100\% .
\end{equation}

We report two levels of success rate, corresponding to loose and strict cutoffs. The loose criterion is defined as
\begin{equation}
\mathcal{C}_{\mathrm{loose}}:
\quad
\mathrm{QED} \ge 0.5,\quad
\mathrm{SAS} \le 6,\quad
\mathrm{Affinity} \le -7.0 .
\end{equation}
This setting represents a relatively accessible drug-like region. The QED threshold of $0.5$ requires molecules to show a reasonable level of drug-likeness. The SAS threshold of $6$ removes molecules that are difficult to synthesize while still allowing moderate synthetic complexity. The Vina threshold of $-7.0$ selects molecules with reasonably strong predicted binding affinity. In addition, the relatively relaxed thresholds help retain a broader set of potentially promising candidates, reducing the risk of prematurely discarding molecules that may be further optimized. Therefore, the loose success rate reflects whether a method can generally guide generation toward molecules with acceptable drug-like properties.

The strict criterion is defined as
\begin{equation}
\mathcal{C}_{\mathrm{strict}}:
\quad
\mathrm{QED} \ge 0.6,\quad
\mathrm{SAS} \le 4,\quad
\mathrm{Affinity} \le -7.5 .
\end{equation}
This setting defines a more selective region corresponding to higher-quality molecular candidates. Compared with the loose criterion, it requires stronger drug-likeness, better synthetic accessibility, and stronger predicted binding affinity. The strict success rate is therefore more challenging and more discriminative. A method with a high strict success rate is able to generate molecules that satisfy multiple desirable properties under more demanding constraints, and thus indicates a higher likelihood of identifying practically viable, lead-like candidates for downstream drug discovery.

Using both loose and strict success rates provides a more comprehensive evaluation. The loose success rate measures whether the model can guide generated molecules into a broadly acceptable property region, while the strict success rate evaluates whether the model has the capability to produce high-quality candidate molecules. A high loose success rate combined with a low strict success rate indicates that the model can improve molecular properties but may struggle to reach top-tier candidates. In contrast, a high strict success rate suggests that the model has strong potential for de novo molecular design applications.

\subsubsection{Additional Structural and Geometric metrics}

Although the above metrics provide a comprehensive assessment of molecular quality from chemical and property-oriented perspectives, they do not explicitly evaluate whether the generated 3D molecules are structurally coherent and geometrically realistic. For example, a molecule may be valid after sanitization but still contain disconnected fragments, distorted local geometry, or physically implausible conformations. Therefore, we further evaluate generated molecules using additional structural and geometry-level metrics, including Connected Fraction and a set of PoseBusters-based metrics.

\paragraph{Connected Fraction.}

Connected Fraction measures the proportion of generated molecules that form a single connected molecular graph. This metric evaluates whether a generated sample corresponds to one coherent molecule rather than multiple disconnected fragments. In our implementation, molecular fragments are identified using RDKit, and a molecule is counted as connected if it contains only one component. Formally,
\begin{equation}
\mathrm{Connected\ Fraction}
=
\frac{
\left|\left\{ m \in \mathcal{M}_{\mathrm{gen}} \mid m \text{ is connected} \right\}\right|
}{
\left|\mathcal{M}_{\mathrm{gen}}\right|
}
\times 100\% .
\end{equation}
The value ranges from $0\%$ to $100\%$, where a higher value indicates that the model more reliably generates complete molecular graphs instead of fragmented structures.

Beyond graph-level connectivity, we use \textbf{PoseBusters} to assess the geometric and physical plausibility of generated 3D molecules. PoseBusters performs a series of structure-quality checks designed to detect common failure modes in generated or predicted molecular conformations, such as unrealistic bond lengths, invalid bond angles, steric clashes, improper ring conformations, and incorrect double-bond geometry. These metrics are important because downstream tasks such as docking, binding analysis, and molecular simulation require not only chemically valid molecular graphs but also physically reasonable 3D conformations. Therefore, PoseBusters-based metrics provide a complementary evaluation of whether generated molecules are suitable for structure-based drug discovery applications.

\paragraph{Bond Lengths Valid.}

Bond Lengths Valid measures the proportion of molecules whose bond lengths fall within chemically acceptable ranges:
\begin{equation}
\mathrm{Bond\ Lengths\ Valid}
=
\frac{
\left|\left\{ m \in \mathcal{M}_{\mathrm{gen}} \mid \text{all bond lengths in } m \text{ are valid} \right\}\right|
}{
\left|\mathcal{M}_{\mathrm{gen}}\right|
}
\times 100\% .
\end{equation}
A higher value indicates that the generated 3D structures preserve realistic bond distances.

\paragraph{Bond Angles Valid.}

Bond Angles Valid evaluates whether the bond angles in generated molecules are chemically plausible:
\begin{equation}
\mathrm{Bond\ Angles\ Valid}
=
\frac{
\left|\left\{ m \in \mathcal{M}_{\mathrm{gen}} \mid \text{all bond angles in } m \text{ are valid} \right\}\right|
}{
\left|\mathcal{M}_{\mathrm{gen}}\right|
}
\times 100\% .
\end{equation}
A higher value reflects more realistic local molecular geometry.

\paragraph{Steric Clash Free.}

Steric Clash Free measures the proportion of molecules without severe atomic overlaps:
\begin{equation}
\mathrm{Steric\ Clash\ Free}
=
\frac{
\left|\left\{ m \in \mathcal{M}_{\mathrm{gen}} \mid m \text{ has no steric clashes} \right\}\right|
}{
\left|\mathcal{M}_{\mathrm{gen}}\right|
}
\times 100\% .
\end{equation}
A higher value indicates that atoms are arranged in physically feasible 3D positions.

\paragraph{Non-Aromatic Ring Non-Flatness.}

Non-Aromatic Ring Non-Flatness evaluates whether non-aromatic rings adopt appropriate non-planar conformations:
\begin{equation}
\mathrm{Non\text{-}Aromatic\ Ring\ Non\text{-}Flatness}
=
\frac{
\left|\left\{ m \in \mathcal{M}_{\mathrm{gen}} \mid \text{non-aromatic rings in } m \text{ are non-flat} \right\}\right|
}{
\left|\mathcal{M}_{\mathrm{gen}}\right|
}
\times 100\% .
\end{equation}
A higher value suggests more realistic ring conformations, since non-aromatic rings are generally expected to deviate from perfect planarity.

\paragraph{Double Bond Flatness.}

Double Bond Flatness measures whether double-bond regions maintain the expected planar geometry:
\begin{equation}
\mathrm{Double\ Bond\ Flatness}
=
\frac{
\left|\left\{ m \in \mathcal{M}_{\mathrm{gen}} \mid \text{double bonds in } m \text{ are planar} \right\}\right|
}{
\left|\mathcal{M}_{\mathrm{gen}}\right|
}
\times 100\% .
\end{equation}
A higher value indicates better preservation of double-bond geometric constraints.

\paragraph{PoseBusters Valid.}

PoseBusters Valid measures the proportion of generated molecules that pass all PoseBusters checks simultaneously:
\begin{equation}
\mathrm{PoseBusters\ Valid}
=
\frac{
\left|\left\{ m \in \mathcal{M}_{\mathrm{gen}} \mid m \text{ passes all PoseBusters checks} \right\}\right|
}{
\left|\mathcal{M}_{\mathrm{gen}}\right|
}
\times 100\% .
\end{equation}
This metric summarizes the overall geometric validity of generated molecules. A high PoseBusters Valid score indicates that the generated molecules are not only chemically valid but also geometrically and physically plausible, which is important for downstream docking and structure-based evaluation.

\subsection{ADMET Analysis and MD Simulation}
\label{appendix:ADMET_Analysis_and_MD_Simulation}

To further evaluate whether the generated molecules exhibit realistic pharmacological properties and stable binding behavior under biologically relevant conditions, we perform ADMET property prediction and MD simulations on the top-ranked candidates. ADMET analysis is used to assess pharmacokinetic properties and potential toxicity, providing an estimate of drug-likeness beyond static structural and docking-based evaluations. MD simulations are used to evaluate the dynamic stability of protein–ligand complexes under realistic physical conditions, verifying whether the predicted binding interactions can be maintained over time.

\paragraph{ADMET.}

ADMET properties are predicted using ADMET-AI~\cite{ADMET-AI}, a graph neural network-based platform trained on large-scale experimental datasets. Although the tool provides predictions for a wide range of endpoints, we focus on a curated subset of 14 key properties that are commonly used in early-stage drug discovery. These properties cover the five major ADMET categories: absorption (e.g., human intestinal absorption, Caco-2 permeability), distribution (e.g., volume of distribution, blood-brain barrier permeability), metabolism (e.g., CYP450 isoform inhibition including CYP3A4, CYP2D6, and CYP2C9), excretion (e.g., renal clearance), and toxicity (e.g., hERG inhibition, Ames mutagenicity, hepatotoxicity, and LD50). 

This subset is selected due to its strong empirical relevance to downstream clinical outcomes and its widespread use as early-stage screening criteria, while avoiding redundancy and reducing the influence of less reliable or weakly predictive endpoints. Each molecule is evaluated both at the individual property level and as an integrated ADMET profile to identify potential liabilities. Molecules that exhibit favorable and balanced performance across all categories are considered pharmacokinetically viable, whereas those with significant deficiencies in critical aspects, such as metabolism or toxicity, are excluded. This analysis serves as an effective early-stage screening step for prioritizing candidates and filtering out compounds with unfavorable pharmacokinetic or safety profiles.

\paragraph{MD simulation.}

MD simulations are performed using Amber. System preparation is carried out using AmberTools~\cite{ambertools}, and all simulations are executed using the GPU-accelerated \texttt{pmemd.cuda} engine.

Protein parameters are assigned using the ff14SB~\cite{ff14sb} force field, while ligand parameters are generated using the General Amber Force Field (GAFF)~\cite{gaff}. Ligand partial atomic charges are assigned using the AM1-BCC method implemented in Antechamber. Each protein-ligand complex is solvated in a rectangular box of OPC water molecules with a minimum padding distance of \(10~\text{\AA}\) from the solute in all directions and neutralized by the addition of counterions. Periodic boundary conditions are applied throughout the simulations.

Prior to production simulations, each system is energy-minimized and subsequently equilibrated using a two-stage protocol consisting of constant-volume (NVT) and constant-pressure (NPT) phases. During equilibration, solute heavy atoms are restrained with a harmonic potential of \(10~\mathrm{kcal\ mol^{-1}}\,\text{\AA}^{-2}\), while temperature is gradually raised to \(300~\mathrm{K}\) using a Langevin thermostat. Pressure is maintained at \(1~\mathrm{atm}\).

Production MD simulations are carried out under the NPT ensemble at \(300~\mathrm{K}\) and \(1~\mathrm{atm}\) using a timestep of \(2~\mathrm{fs}\), for a total simulation time of \(10~\mathrm{ns}\) (\(5{,}000{,}000\) steps). All bonds involving hydrogen atoms are constrained using the SHAKE algorithm. Long-range electrostatics are treated with the Particle Mesh Ewald (PME) method, and a \(10~\text{\AA}\) cutoff is applied for nonbonded interactions.

These simulation protocols ensure gradual relaxation of the system, proper solvent equilibration, and stable production dynamics while maintaining computational efficiency.

The stability of each protein-ligand complex is assessed using the RMSD of protein backbone atoms over time, computed relative to the initial protein structure. RMSD trajectories are used to evaluate structural stability and equilibration behavior. Stable systems are characterized by convergence of RMSD values followed by fluctuations around a stable plateau, indicating equilibration of the protein structure during the simulation. In contrast, large fluctuations or continuous drift suggest unstable binding or structural inconsistency. This analysis provides preliminary evidence of whether the generated molecules can maintain structurally stable interactions during the MD simulation timescale.

\subsection{Computing Time and Resources}
\label{appendix:Computing_Time_and_Resources}
Experiments are conducted primarily on NVIDIA A100 GPUs with 80 GB of memory, as well as NVIDIA L40 GPUs. Each training episode consists of molecule generation, property evaluation, and model update. On an A100 GPU, optimizing the flow matching model pretrained on GEOM-Drugs takes approximately 2 minutes per episode, while optimizing the model pretrained on QM9 takes approximately 1 minute per episode.

\paragraph{QED and SAS evaluation time.}
We measured the time required to evaluate QED and SAS for sets of generated molecules, most of which contain 20--40 heavy atoms. The runtimes are summarized in Table~\ref{tab:qed_sas_runtime}.

\begin{table}[htbp]
\centering
\caption{Runtime for online QED and SAS evaluation.}
\label{tab:qed_sas_runtime}
\begin{tabular}{rrrr}
\toprule
Number of molecules & QED (s) & SAS (s) & Total (s) \\
\midrule
1,000 & 1.03 & 0.20 & 1.23 \\
10,000 & 9.06 & 1.99 & 11.05 \\
\bottomrule
\end{tabular}
\end{table}

\section{Additional Experiments, Results and Discussions}
\label{appendix:Additional_Experiments_Results_and_Discussions}
\subsection{Large-Scale Generation Evaluation}

We further evaluated 10,000 generated molecules for each target. Our method achieved Loose and Strict success rates of 87.64\% and 70.72\% on EGFR and 7.47\% and 0.76\% on Pin1, while maintaining high validity and VUN (Table~\ref{tab:large_generation}). These results are consistent with the improvements observed in the smaller-scale evaluation.

\begin{table}[htbp]
\centering
\small
\caption{
Large-scale generation evaluation (10,000 molecules).
}
\label{tab:large_generation}
\resizebox{\columnwidth}{!}{
\begin{tabular}{lcccc}
\toprule
&
\multicolumn{2}{c}{\textbf{EGFR (PDB: 3W2S)}} &
\multicolumn{2}{c}{\textbf{Pin1 (PDB: 3IKD)}} \\
\cmidrule(lr){2-3}
\cmidrule(lr){4-5}
\textbf{Metrics} &
\textbf{Base Model} &
\textbf{Ours} &
\textbf{Base Model} &
\textbf{Ours} \\
\midrule

Validity (\%) $\uparrow$ &
98.99 &
99.64 &
98.99 &
99.52 \\

Uniqueness (\%) $\uparrow$ &
99.89 &
99.56 &
99.89 &
99.94 \\

Novelty (\%) $\uparrow$ &
100.00 &
100.00 &
100.00 &
100.00 \\

Validity $\times$ Uniqueness $\times$ Novelty (\%) $\uparrow$ &
98.88 &
99.20 &
98.88 &
99.46 \\

Atom Stability (\%) $\uparrow$ &
99.99 &
100.00 &
99.99 &
99.99 \\

Molecule Stability (\%) $\uparrow$ &
99.75 &
99.83 &
99.75 &
99.54 \\

Success Rate (Loose, \%) $\uparrow$ &
69.35 &
87.64 &
5.07 &
7.47 \\

Success Rate (Strict, \%) $\uparrow$ &
46.30 &
70.72 &
0.38 &
0.76 \\

\bottomrule
\end{tabular}
}
\end{table}

\subsection{PoseBuster Evaluation}

As described in Appendix~\ref{appendix:Performance_Metrics}, we also use PoseBusters to evaluate the generated molecules. Overall PoseBusters validity increases from 92.10\% to 94.45\% on EGFR and from 92.10\% to 92.70\% on Pin1 (Table~\ref{tab:posebuster_results}), indicating that the generated molecules maintain high structural plausibility.

\begin{table}[htbp]
\centering
\small
\caption{
PoseBusters-based evaluation.
}
\label{tab:posebuster_results}
\resizebox{\columnwidth}{!}{
\begin{tabular}{lcccc}
\toprule
&
\multicolumn{2}{c}{\textbf{EGFR (PDB: 3W2S)}} &
\multicolumn{2}{c}{\textbf{Pin1 (PDB: 3IKD)}} \\
\cmidrule(lr){2-3}
\cmidrule(lr){4-5}
\textbf{Metrics} &
\textbf{Base Model} &
\textbf{Ours} &
\textbf{Base Model} &
\textbf{Ours} \\
\midrule

Connected Fraction (\%) $\uparrow$ &
99.12 &
99.71 &
99.12 &
99.25 \\

Bond Lengths Valid (\%) $\uparrow$ &
99.99 &
100.00 &
99.99 &
99.98 \\

Bond Angles Valid (\%) $\uparrow$ &
100.00 &
100.00 &
100.00 &
99.99 \\

Steric Clash Free (\%) $\uparrow$ &
94.02 &
95.18 &
94.02 &
94.01 \\

Non-Aromatic Ring Non-Flatness (\%) $\uparrow$ &
99.67 &
99.70 &
99.67 &
99.72 \\

Double Bond Flatness (\%) $\uparrow$ &
99.36 &
99.93 &
99.36 &
99.42 \\

PoseBusters Valid (\%) $\uparrow$ &
92.10 &
94.45 &
92.10 &
92.70 \\

\bottomrule
\end{tabular}
}
\end{table}

\subsection{Additional Sensitivity Analysis}

\begin{figure}[htbp]
\centering
\includegraphics[width=0.99\linewidth]{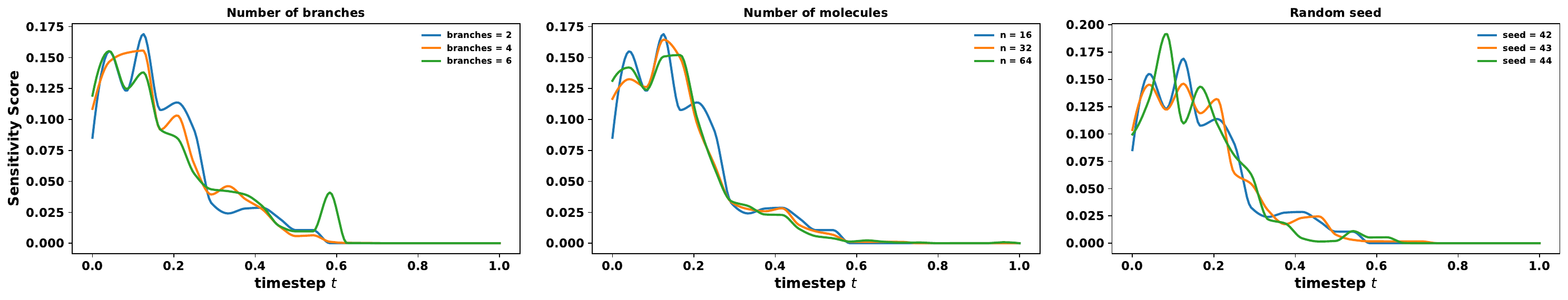}
\caption{Temporal selection ablation. The analyses of branches, number of probes, and seeds use SAS as the target property.}
\label{fig:sas_sensitivity_ablation}
\end{figure}

In addition to using QED as the target property to analyze different numbers of branches, probe molecules, and random seeds, we further repeat the same analyses using SAS as the target property to verify the robustness and generality of the temporal selection behavior across different molecular objectives. The results are shown in Figure~\ref{fig:sas_sensitivity_ablation}, demonstrating consistent temporal sensitivity patterns under different experimental settings.

\subsection{Generated Top Candidate}

\begin{figure}[htbp]
\centering
\includegraphics[width=0.99\linewidth]{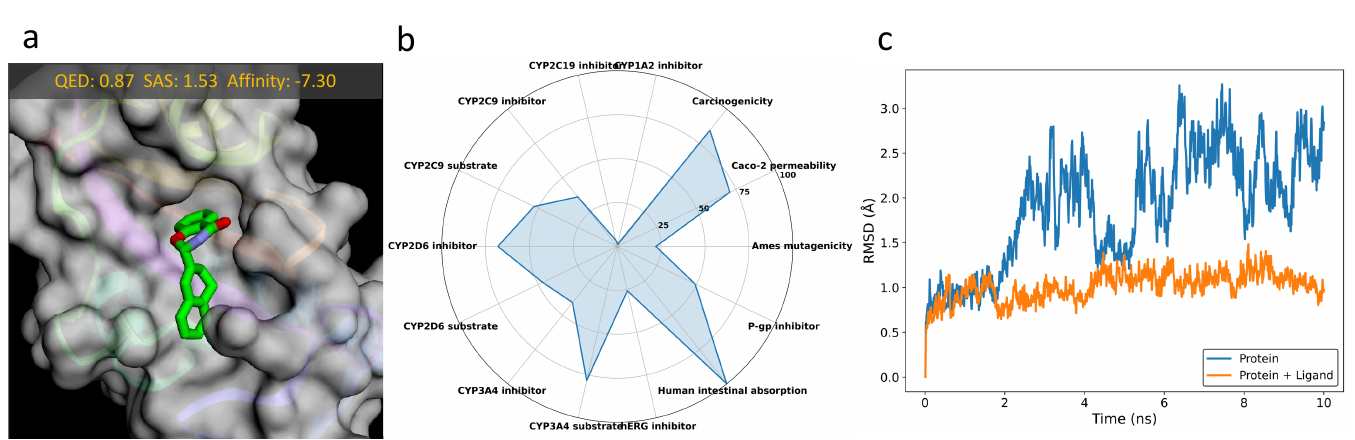}
\caption{A top candidate for Pin1 (3IKD). The three subfigures in each row represent docking visualization, ADMET analysis (farther from center is better), and MD simulation RMSD trajectories (lower is more stable) from left to right, respectively.}
\label{fig:another_pin1_candidate}
\end{figure}

Figure~\ref{fig:another_pin1_candidate} shows downstream analyses for a generated candidate molecule for Pin1.

\subsection{Generalization Experiment}
\label{appendix:Generalization_Experiment}

To further evaluate generalization, experiments are also conducted on the QM9 dataset. In this setting, instead of drug-related properties, five physical properties are considered, namely HOMO, LUMO, gap, $\mu$, and $\alpha$, measured using DimeNet++~\cite{dimenetpp_2020}, to assess the effectiveness of the proposed method under different property types and a larger set of optimization objectives. The results in Table~\ref{tab:qm9_results} demonstrate that the proposed framework remains effective across different datasets, molecular properties, and optimization objectives, indicating the generality of the method under diverse molecular design settings.

\begin{table}[htbp]
\centering
\small
\caption{
Performance on the QM9 dataset.
}
\label{tab:qm9_results}
\begin{tabular}{lcc}
\toprule
&
\multicolumn{2}{c}{\textbf{QM9}} \\
\cmidrule(lr){2-3}
\textbf{Metrics} &
\textbf{Base Model} &
\textbf{Ours} \\
\midrule

Validity (\%) $\uparrow$ &
99.81 &
99.83 \\

Uniqueness (\%) $\uparrow$ &
92.54 &
91.97 \\

Novelty (\%) $\uparrow$ &
99.99 &
99.86 \\

Validity $\times$ Uniqueness $\times$ Novelty (\%) $\uparrow$ &
92.35 &
91.68 \\

Atom Stability (\%) $\uparrow$ &
99.99 &
100.00 \\

Molecule Stability (\%) $\uparrow$ &
99.82 &
99.94 \\

Success Rate (\%) $\uparrow$ &
23.74 &
25.86 \\

Connected Fraction (\%) $\uparrow$ &
99.97 &
99.98 \\

Bond Lengths Valid (\%) $\uparrow$ &
99.98 &
99.97 \\

Bond Angles Valid (\%) $\uparrow$ &
100.00 &
100.00 \\

Steric Clash Free (\%) $\uparrow$ &
99.96 &
100.00 \\

Non-Aromatic Ring Non-Flatness (\%) $\uparrow$ &
99.61 &
99.72 \\

Double Bond Flatness (\%) $\uparrow$ &
99.58 &
99.59 \\

PoseBusters Valid (\%) $\uparrow$ &
98.67 &
98.79 \\

\bottomrule
\end{tabular}

\vspace{2pt}

\raggedright
\footnotesize
\textbf{Note.}
For QM9, the success rate is defined as the fraction of generated molecules whose five target properties simultaneously fall within the predefined target ranges:
HOMO $\in [-6.5,-5.5]$ eV,
LUMO $\in [-1.5,-0.5]$ eV,
gap $\in [4.0,6.0]$ eV,
$\mu \in [2.0,5.0]$ Debye,
and $\alpha \in [60,120]\,a_0^3$. Test based on 10,000 generated molecules.
\end{table}

\section{Limitations and Broader Impact}

\paragraph{Limitations.} Molecular evaluation in this work mainly relies on computational metrics and in silico simulations, which may not fully reflect experimental biological activity or real-world synthesizability. Future work will involve wet-lab synthesis and experimental validation of top candidate molecules.

\paragraph{Broader Impact.}
This work presents a generative framework that may accelerate early-stage drug discovery by effectively exploring chemical space and optimizing multiple drug-relevant properties. Beyond drug discovery, the approach may also benefit related applications such as materials science, catalyst design, and molecular engineering. At the same time, powerful molecular generative models may potentially be misused to design harmful compounds. We therefore support the responsible use of such technologies within appropriate regulatory and safety frameworks.

\section{Source Code}
The source code is available at \url{https://github.com/Kyle4490/TemporalFM}.

\newpage
\clearpage
\section*{NeurIPS Paper Checklist}

\begin{enumerate}

\item {\bf Claims}
    \item[] Question: Do the main claims made in the abstract and introduction accurately reflect the paper's contributions and scope?
    \item[] Answer: \answerYes{} %
    \item[] Justification: The abstract and introduction state the main contributions, namely a multi-objective reinforcement learning framework for adaptively optimizing continuous and discrete flow matching models for 3D molecular generation. The claims are supported by the methodological formulation and empirical results. The limitations of the method are acknowledged in the final section. Moreover, while the long-term motivation of aiding real-world drug discovery is mentioned, it is clearly separated from the achieved contributions. Therefore, the claims accurately reflect the paper’s scope and results.

\item {\bf Limitations}
    \item[] Question: Does the paper discuss the limitations of the work performed by the authors?
    \item[] Answer: \answerYes{} %
    \item[] Justification: The limitations of our work are discussed in Appendix E.

\item {\bf Theory assumptions and proofs}
    \item[] Question: For each theoretical result, does the paper provide the full set of assumptions and a complete (and correct) proof?
    \item[] Answer: \answerYes{} %
    \item[] Justification: The paper provides theoretical results and proofs in the method section and the appendix.

    \item {\bf Experimental result reproducibility}
    \item[] Question: Does the paper fully disclose all the information needed to reproduce the main experimental results of the paper to the extent that it affects the main claims and/or conclusions of the paper (regardless of whether the code and data are provided or not)?
    \item[] Answer: \answerYes{} %
    \item[] Justification: All experimental settings, including data splits, model backbones, hyperparameters, and evaluation protocols, are described in the experimental section and appendix.

\item {\bf Open access to data and code}
    \item[] Question: Does the paper provide open access to the data and code, with sufficient instructions to faithfully reproduce the main experimental results, as described in supplemental material?
    \item[] Answer: \answerYes{} %
    \item[] Justification: The datasets used in our work are publicly available, with access links provided in the appendix. The source code will be released upon acceptance of the manuscript.

\item {\bf Experimental setting/details}
    \item[] Question: Does the paper specify all the training and test details (e.g., data splits, hyperparameters, how they were chosen, type of optimizer) necessary to understand the results?
    \item[] Answer: \answerYes{} %
    \item[] Justification: All experimental details, including dataset descriptions, data splits, model architectures, optimization procedures, hyperparameter configurations, and evaluation protocols, are clearly described in the experimental section and the appendix.

\item {\bf Experiment statistical significance}
    \item[] Question: Does the paper report error bars suitably and correctly defined or other appropriate information about the statistical significance of the experiments?
    \item[] Answer: \answerYes{} %
    \item[] Justification: We report standard deviation across multiple runs for all key experiments.

\item {\bf Experiments compute resources}
    \item[] Question: For each experiment, does the paper provide sufficient information on the computer resources (type of compute workers, memory, time of execution) needed to reproduce the experiments?
    \item[] Answer: \answerYes{} %
    \item[] Justification: We mention the computational resources used to conduct the experiments in the experimental section and the appendix.
    
\item {\bf Code of ethics}
    \item[] Question: Does the research conducted in the paper conform, in every respect, with the NeurIPS Code of Ethics \url{https://neurips.cc/public/EthicsGuidelines}?
    \item[] Answer: \answerYes{} %
    \item[] Justification: We have reviewed the NeurIPS Code of Ethics, and our research follows the guidelines. Our work does not involve human subjects or sensitive data.

\item {\bf Broader impacts}
    \item[] Question: Does the paper discuss both potential positive societal impacts and negative societal impacts of the work performed?
    \item[] Answer: \answerYes{} %
    \item[] Justification: Broader impacts are discussed in Appendix E.
    
\item {\bf Safeguards}
    \item[] Question: Does the paper describe safeguards that have been put in place for responsible release of data or models that have a high risk for misuse (e.g., pre-trained language models, image generators, or scraped datasets)?
    \item[] Answer: \answerNA{} %
    \item[] Justification: Our work does not involve models or datasets that pose a high risk of misuse.

\item {\bf Licenses for existing assets}
    \item[] Question: Are the creators or original owners of assets (e.g., code, data, models), used in the paper, properly credited and are the license and terms of use explicitly mentioned and properly respected?
    \item[] Answer: \answerYes{} %
    \item[] Justification: All external tools (e.g., RDKit, AutoDock Vina) and models are cited and used under their respective licenses.

\item {\bf New assets}
    \item[] Question: Are new assets introduced in the paper well documented and is the documentation provided alongside the assets?
    \item[] Answer: \answerYes{} %
    \item[] Justification: We introduce new code and models, which are documented and will be released with usage instructions upon acceptance.

\item {\bf Crowdsourcing and research with human subjects}
    \item[] Question: For crowdsourcing experiments and research with human subjects, does the paper include the full text of instructions given to participants and screenshots, if applicable, as well as details about compensation (if any)? 
    \item[] Answer: \answerNA{} %
    \item[] Justification: This work does not involve crowdsourcing or research involving human subjects.

\item {\bf Institutional review board (IRB) approvals or equivalent for research with human subjects}
    \item[] Question: Does the paper describe potential risks incurred by study participants, whether such risks were disclosed to the subjects, and whether Institutional Review Board (IRB) approvals (or an equivalent approval/review based on the requirements of your country or institution) were obtained?
    \item[] Answer: \answerNA{} %
    \item[] Justification: This work does not involve research involving human subjects and therefore does not require IRB approval.

\item {\bf Declaration of LLM usage}
    \item[] Question: Does the paper describe the usage of LLMs if it is an important, original, or non-standard component of the core methods in this research? Note that if the LLM is used only for writing, editing, or formatting purposes and does \emph{not} impact the core methodology, scientific rigor, or originality of the research, declaration is not required.
    \item[] Answer: \answerNA{} %
    \item[] Justification: LLMs are not used as a component of the core methods in this research.

\end{enumerate}

\end{document}